\documentclass[11pt]{article}

\usepackage[final]{acl}

\usepackage{times}
\usepackage{latexsym}
\usepackage[T1]{fontenc}
\usepackage[utf8]{inputenc}
\usepackage{microtype}
\usepackage{inconsolata}
\usepackage{graphicx}

\usepackage{subcaption}
\usepackage{booktabs}
\usepackage{multirow}
\usepackage{colortbl}
\usepackage{float}
\usepackage{algorithm}
\usepackage{algorithmic}
\newcommand{\FUNCTION}[1]{\STATE \textbf{Function} #1}
\newcommand{\ENDFUNCTION}{\STATE \textbf{end function}}
\usepackage{amsmath,amssymb,mathtools}
\usepackage{tikz}
\usetikzlibrary{arrows.meta, positioning, calc, fit,
                shapes.geometric, decorations.pathreplacing,
                patterns, shadows.blur, backgrounds}
\usepackage{tcolorbox}
\usepackage{enumitem}
\usepackage[capitalize,noabbrev]{cleveref}
\usepackage{verbatim}

\definecolor{hueF}{HTML}{4E6B52}   
\definecolor{hueA}{HTML}{2E4A6B}   

\providecommand{\nolinenumbers}{}
\providecommand{\linenumbers}{}

\BeforeBeginEnvironment{tcolorbox}{\par\nolinenumbers}
\AfterEndEnvironment{tcolorbox}{\linenumbers}

\BeforeBeginEnvironment{table}{\par\nolinenumbers}
\AfterEndEnvironment{table}{\linenumbers}
\BeforeBeginEnvironment{table*}{\par\nolinenumbers}
\AfterEndEnvironment{table*}{\linenumbers}

\title{SemEnrich: Self-Supervised Semantic Enrichment of Radiology Reports
       for Vision-Language Learning}

\author{Halil Ibrahim Gulluk \\
  Department of Electrical Engineering \\
  Stanford University \\
  \texttt{gulluk@stanford.edu} \\\And
  Olivier Gevaert \\
  Department of Biomedical Data Science \\
  Stanford University \\
  \texttt{ogevaert@stanford.edu}}

\begin{document}
\maketitle

\begin{abstract}
    Medical vision--language datasets are limited in size and biased toward abnormal findings: clinicians document abnormalities in detail but frequently omit normal observations because they are considered uninformative for the patient's condition.
    We propose a self-supervised enrichment method that leverages semantic clustering of report sentences and augments the training reports with normal observations drawn from co-occurring clusters in the same dataset.
    Our approach yields consistent gains in supervised fine-tuning: average improvements of $+5.63\%$ COMET, $+3.04\%$ BERTScore, $+7.40\%$ Sentence-BLEU, $+5.30\%$ CheXbert-F1, and $+7.47\%$ RadGraph-F1 across five language model backbones.
    Ablation studies confirm that improvements stem from semantic clustering rather than random augmentation.
    We further introduce a cluster-aware reward for GRPO training that complements exact-match rewards and yields additional gains: $+2.78\%$ COMET, $+3.14\%$ BERTScore, and $+14.21\%$ ROUGE-L.
    Code is available at \href{https://github.com/igulluk/SemEnrich}{https://github.com/igulluk/SemEnrich}.
\end{abstract}

\section{Introduction}



Vision--language models trained on large-scale web data have made rapid progress on image captioning, visual question answering, and text-to-image generation \cite{radford2021learning,li2023blip,jia2021scaling}. The dominant lesson from this line of work is simple: the larger and richer the training data, the better the downstream performance.

The medical domain, however, breaks this scaling assumption. Public medical image--text corpora are small, expensive to curate, and biased in ways general-domain data is not. Many medical visual question answering benchmarks contain only coarse prompts (\textit{e.g.}, ``which organ is presented'' or ``which imaging modality is shown''), which are insufficient to evaluate models intended for real clinical use \cite{lau2018dataset,zhang2024development}. Building new medical datasets is one response; improving existing ones is the complementary, lower-cost path that we pursue in this work.

To make this concrete, we focus on chest X-ray reports, which contain two main sections: \emph{findings} and \emph{impressions}. The findings section is a list of observations made by the radiologist from the image; the impression is the radiologist's final conclusion, often informed by patient history and clinical indication in addition to the image itself. Because the impression depends on information outside the image, we target the findings section, where a large vision--language model (LVLM) can in principle ground every statement in the visual input.

The findings section also has a structural property that ordinary text data does not: its sentences are largely independent. A radiologist may write the same set of observations in any order without changing the meaning of the report. This order invariance lets us treat the findings as an \emph{unordered set} of observations and, after embedding each sentence, group them into semantic clusters and reason about which clusters can co-occur.

A second property of these reports motivates our enrichment. When writing a report, clinicians focus on abnormalities and tend to omit normal observations that they consider uninformative for the patient's condition, so the training data systematically under-represents normal anatomy. Crucially, the two properties above let us repair this bias without external supervision: because findings are an exchangeable set, a normal observation that has been omitted can be reinserted as long as it is consistent with what the report already states.

\paragraph{Our approach.}
We introduce \textbf{SemEnrich}, a self-supervised pipeline that exploits both properties. We embed every finding sentence in the training corpus with a sentence transformer and group the embeddings into semantic clusters. Each cluster is assigned a \emph{polarity} (normal or abnormal), and two clusters are deemed \emph{connected} when they already co-occur in some real report. For a given patient we then add only normal clusters that are mutually connected to every cluster already present -- a \emph{valid enrichment} -- so that the inserted sentences are plausibly omitted normal findings rather than fabricated content. The same clustering is reused as a reward signal for GRPO post-training, where a cluster-level F1 reward complements the standard exact-match reward. We apply this type of self-supervision to the training set and the test set is never modified, so all reported gains reflect a richer training signal rather than easier targets.



\begin{figure*}[t]
\centering
\input{figs/paper_style}
\resizebox{\textwidth}{!}{%
\begin{tikzpicture}[font=\sffamily, x=1cm, y=1cm]


\headerstripbox{0.00}{8.00}{14.60}{8.50}{hueA}
  {Phase 1 -- Offline preprocessing}
  {embed every training sentence, cluster, label polarity, build connectivity}

\node[font=\scriptsize\bfseries, color=hueA] at (1.25, 7.55) {Training corpus};

\foreach \i/\dx/\dy in {0/0/0, 1/0.12/-0.18, 2/0.24/-0.36} {
  \fill[hueAl!22, rounded corners=2pt]
        (0.40+\dx, 5.50+\dy) rectangle (1.65+\dx, 7.30+\dy);
  \draw[hueA!65, line width=0.5pt, rounded corners=2pt]
        (0.40+\dx, 5.50+\dy) rectangle (1.65+\dx, 7.30+\dy);
  \foreach \k in {0,1,2,3,4} {
    \draw[hueA!55, line width=0.35pt]
      (0.55+\dx, 7.05+\dy-0.30*\k) -- (1.50+\dx, 7.05+\dy-0.30*\k);
  }
}

\node[font=\tiny, color=inkmed] at (1.25, 4.85)
  {184{,}535 finding sentences};

\draw[arw] (2.20, 6.30) -- (3.00, 6.30);

\node[font=\scriptsize\bfseries, color=hueB] at (3.65, 7.55) {Sentence embedder};

\begin{scope}[shift={(3.10, 5.40)}]
  \trapezoidNarr{hueB}{MiniLM}{text}{$\mathbb{R}^{384}$}
\end{scope}

\node[font=\tiny, color=inkmed] at (3.65, 4.85) {pretrained, frozen};

\draw[arw] (4.30, 6.30) -- (5.10, 6.30);

\node[font=\scriptsize\bfseries, color=hueC] at (6.30, 7.55) {Sentence clusters};

\fill[hueC!20] (5.65, 6.60) circle (0.30);
\draw[hueC!75, line width=0.6pt] (5.65, 6.60) circle (0.30);
\node[font=\tiny\bfseries, color=hueC!40!black] at (5.65, 6.60) {$c_1$};

\fill[hueC!20] (6.65, 6.75) circle (0.28);
\draw[hueC!75, line width=0.6pt] (6.65, 6.75) circle (0.28);
\node[font=\tiny\bfseries, color=hueC!40!black] at (6.65, 6.75) {$c_2$};

\fill[hueC!20] (5.95, 5.80) circle (0.28);
\draw[hueC!75, line width=0.6pt] (5.95, 5.80) circle (0.28);
\node[font=\tiny\bfseries, color=hueC!40!black] at (5.95, 5.80) {$c_3$};

\fill[hueC!20] (6.85, 5.85) circle (0.26);
\draw[hueC!75, line width=0.6pt] (6.85, 5.85) circle (0.26);
\node[font=\tiny\bfseries, color=hueC!40!black] at (6.85, 5.85) {$c_4$};

\fill[hueC!20] (7.15, 6.40) circle (0.24);
\draw[hueC!75, line width=0.6pt] (7.15, 6.40) circle (0.24);
\node[font=\tiny\bfseries, color=hueC!40!black] at (7.15, 6.40) {$c_5$};

\node[font=\tiny, color=inkmed] at (6.30, 4.85)
  {HDBSCAN, $K{=}3{,}216$};

\draw[arw] (7.55, 6.30) -- (8.35, 6.30);

\node[font=\scriptsize\bfseries, color=hueA] at (9.20, 7.55) {Cluster polarity};

\node[draw, rounded corners=4pt, fill=hueF!18, draw=hueF!70,
      minimum width=1.30cm, minimum height=0.42cm, inner sep=2pt,
      font=\scriptsize\sffamily\color{hueF!40!black}] at (9.20, 6.65) {Normal};
\node[draw, rounded corners=4pt, fill=hueE!18, draw=hueE!70,
      minimum width=1.30cm, minimum height=0.42cm, inner sep=2pt,
      font=\scriptsize\sffamily\color{hueE!40!black}] at (9.20, 6.05) {Abnormal};
\node[font=\tiny, color=inkmed] at (9.20, 5.55) {LLM judge};

\node[font=\tiny, color=inkmed] at (9.20, 4.85)
  {$h_S\!: c \!\to\! \{\pm 1\}$};

\draw[arw] (9.95, 6.30) -- (10.75, 6.30);

\node[font=\scriptsize\bfseries, color=hueA] at (11.85, 7.55) {Connectivity};

\node[circle, draw=hueF!75, fill=hueF!22, line width=0.5pt,
      minimum size=5mm, inner sep=0pt,
      font=\tiny\color{hueF!40!black}] (g1) at (11.10, 6.55) {$c_1$};
\node[circle, draw=hueE!75, fill=hueE!22, line width=0.5pt,
      minimum size=5mm, inner sep=0pt,
      font=\tiny\color{hueE!40!black}] (g2) at (11.85, 6.90) {$c_2$};
\node[circle, draw=hueE!75, fill=hueE!22, line width=0.5pt,
      minimum size=5mm, inner sep=0pt,
      font=\tiny\color{hueE!40!black}] (g3) at (11.85, 6.15) {$c_3$};
\node[circle, draw=hueF!75, fill=hueF!22, line width=0.5pt,
      minimum size=5mm, inner sep=0pt,
      font=\tiny\color{hueF!40!black}] (g4) at (12.60, 6.55) {$c_4$};
\node[circle, draw=hueF!75, fill=hueF!22, line width=0.5pt,
      minimum size=5mm, inner sep=0pt,
      font=\tiny\color{hueF!40!black}] (g5) at (11.85, 5.55) {$c_5$};
\draw[inkmed!55, line width=0.4pt] (g1)--(g2);
\draw[inkmed!55, line width=0.4pt] (g1)--(g3);
\draw[inkmed!55, line width=0.4pt] (g1)--(g4);
\draw[inkmed!55, line width=0.4pt] (g2)--(g4);
\draw[inkmed!55, line width=0.4pt] (g3)--(g4);
\draw[inkmed!55, line width=0.4pt] (g2)--(g3);
\draw[inkmed!55, line width=0.4pt] (g3)--(g5);
\draw[inkmed!55, line width=0.4pt] (g1)--(g5);
\draw[inkmed!55, line width=0.4pt] (g4)--(g5);

\node[font=\tiny, color=inkmed] at (11.85, 4.85)
  {co-occurrence graph};

\headerstripbox{0.00}{3.40}{14.60}{3.90}{hueB}
  {Phase 2 -- Training-time enrichment}
  {sample valid normal sentences, append to findings, train LVLM (SFT or GRPO)}

\node[font=\scriptsize\bfseries, color=hueA] at (1.47, 3.05) {Original report};

\node[draw=hueA!70, rounded corners=2pt, line width=0.5pt, inner sep=1pt]
      (xray_orig) at (0.90, 1.95) {%
  \includegraphics[width=0.80cm, height=0.80cm]{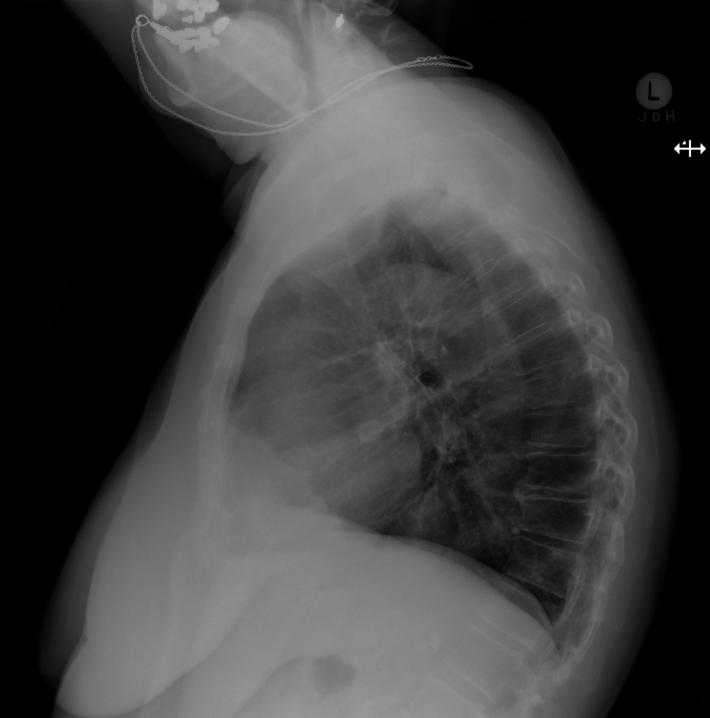}%
};
\node[font=\scriptsize, color=hueA!60!black] at (0.90, 1.30) {$v_j$};

\tokenrow{1.70}{1.20}{4}{hueA}{}{$F_j$}

\node[font=\tiny, color=inkmed] at (1.47, 0.55)
  {abnormalities dominate};

\draw[arw] (2.50, 1.95) -- (3.30, 1.95);

\node[font=\scriptsize\bfseries, color=hueF] at (4.42, 3.05)
  {Sample valid enrichment};

\fill[hueF!12, rounded corners=4pt] (3.42, 1.15) rectangle (5.42, 2.75);
\draw[hueF!65, line width=0.5pt, rounded corners=4pt]
  (3.42, 1.15) rectangle (5.42, 2.75);

\node[font=\tiny, color=hueF!35!black, align=center]
  at (4.42, 2.35) {connected normal\\clusters $F_{cand}$};

\node[circle, draw=hueF!75, fill=hueF!28, line width=0.5pt,
      minimum size=4mm, inner sep=0pt,
      font=\tiny\color{hueF!40!black}] at (3.77, 1.60) {$c_5$};
\node[circle, draw=hueF!75, fill=hueF!28, line width=0.5pt,
      minimum size=4mm, inner sep=0pt,
      font=\tiny\color{hueF!40!black}] at (4.42, 1.60) {$c_8$};
\node[circle, draw=hueF!75, fill=hueF!28, line width=0.5pt,
      minimum size=4mm, inner sep=0pt,
      font=\tiny\color{hueF!40!black}] at (5.07, 1.60) {$c_9$};

\node[sumnode] (plus) at (5.90, 1.95) {\small$+$};

\node[font=\tiny, color=inkmed] at (4.42, 0.55)
  {uses connectivity, polarity};

\draw[arw] (6.37, 1.95) -- (6.80, 1.95);

\node[font=\scriptsize\bfseries, color=hueF] at (8.55, 3.05) {Enriched inputs};

\node[font=\fontsize{32}{36}\selectfont, color=hueF!55!black,
      inner sep=1pt] (lparen) at (6.95, 1.95) {(};

\tokenrow{7.40}{1.20}{7}{hueF}{}{$F_j \cup F_{cand}$}

\node[font=\Large\bfseries, color=inkmed] at (8.75, 1.80) {,};

\node[draw=hueA!70, rounded corners=2pt, line width=0.5pt, inner sep=1pt]
      (xray_lvlm) at (9.35, 1.95) {%
  \includegraphics[width=0.80cm, height=0.80cm]{figs/FindingsExample.png}%
};

\node[font=\fontsize{32}{36}\selectfont, color=hueF!55!black,
      inner sep=1pt] (rparen) at (10.10, 1.95) {)};

\node[font=\tiny, color=inkmed] at (8.55, 0.55)
  {findings $+$ image};

\node[font=\scriptsize\bfseries, color=hueD] at (12.05, 3.05) {LVLM training};

\node[draw, rounded corners=4pt, line width=0.8pt,
      fill=hueD!12, draw=hueD!70,
      minimum width=2.00cm, minimum height=1.30cm, inner sep=2pt,
      align=center]
      (lvlm) at (12.05, 1.95) {%
  {\small\bfseries\sffamily\color{hueD!35!black} LVLM}%
};

\draw[arw] (rparen.east) -- (lvlm.west);

\node[font=\tiny, color=inkmed] at (12.05, 0.55)
  {ViT + LLM, SFT or GRPO};

\draw[lossarw] (lvlm.east) -- ++(0.80, 0);

\node[font=\scriptsize\bfseries, color=emph!75!black] at (14.20, 3.05)
  {Loss / reward};

\node[font=\small, color=emph!75!black]
  at (14.20, 1.95) {$\mathcal{L} \,/\, \mathcal{R}_{cl}$};

\node[font=\tiny, color=inkmed] at (14.20, 0.55)
  {training objective};

\draw[modarw]
  (11.85, 4.65) -- (11.85, 4.30) -- (4.42, 4.30) -- (4.42, 3.20);

\end{tikzpicture}%
}
\caption{\textbf{SemEnrich pipeline.} \emph{Phase 1 (top):} every finding sentence in the training set is embedded by a frozen sentence transformer (MiniLM) and clustered with HDBSCAN. Each cluster is labeled \emph{normal} or \emph{abnormal} by an LLM, and cluster co-occurrence across patients defines a connectivity graph. \emph{Phase 2 (bottom):} for every patient $(v_j, F_j)$ we sample a \emph{valid enrichment}---a set of normal clusters $F_{cand}$ connected to every cluster already in $F_j$---and append one sentence per added cluster to the findings. The enriched findings and the chest X-ray then train an LVLM via SFT or GRPO; in the GRPO setting the same clustering is reused as a cluster-F1 reward $\mathcal{R}_{cl}$.}
\label{fig:pipeline}
\end{figure*}

\section{Related Work}

\paragraph{Medical vision--language models:}
A growing line of work targets medical visual question answering and radiology report generation under limited data. \citet{lin2023pmc} curate a medical image--caption dataset and train a CLIP-style model; \citet{wu2024pmc} train a Llama model for medical multiple-choice question answering. LLaVA-Med \cite{li2023llava} extends LLaVA to medical conversations including images, and R-LLaVA \cite{chen2025r} improves medical VQA with region-of-interest grounding. \citet{tu2024towards} propose a generalist medical model that handles VQA, report generation, and classification within a single architecture.

\paragraph{Improving information extraction from medical data:}
Several approaches focus on richer alignment or retrieval rather than larger data. MMed-RAG \cite{xia2024mmed} introduces a domain-aware multi-modal RAG pipeline with preference fine-tuning. Med-UniC \cite{wan2023med} aligns medical text across English and Spanish to broaden coverage; \citet{boecking2022making} show that combining global and local alignment of visual and text representations improves downstream medical VLM performance.

\paragraph{Datasets and report generation models:}
We use ReXGradient-160K \cite{zhang2025rexgradient}, which contains 160K chest X-ray studies across 109{,}487 patients from 79 medical sites; the rationale for this choice is discussed in Appendix~\ref{sec:dataset_appendix}. Related releases include 3D-RAD \cite{gai20263d} for 3D medical VQA and RadVLM \cite{deperrois2025radvlm} for conversational radiology. For report generation specifically, Flamingo-CXR \cite{tanno2025collaboration} adapts a Flamingo backbone with expert evaluation in the loop, \citet{chen2020generating} propose a memory-driven transformer for radiology report generation, and \citet{alfarghaly2021automated} use conditioned transformers for automated report generation.

In contrast to these methods, SemEnrich does not propose a new dataset or a new architecture. It produces an \emph{enrichment} of any existing chest X-ray dataset, by adding semantically consistent normal findings that the radiologist did not bother to mention, and it can be combined with any of the report-generation backbones above.

\section{Method}

This section describes (i) how we cluster the sentences in the findings section across all training reports, (ii) how we label each cluster as normal or abnormal, and (iii) how we use this structure to enrich the training findings in a self-supervised manner.

\subsection{Sentence Representations and Clustering}

We operate on the findings section of each report, which lists observations made directly from the chest X-ray.

\begin{figure}[ht]
    \centering
    \includegraphics[width=0.5\columnwidth]{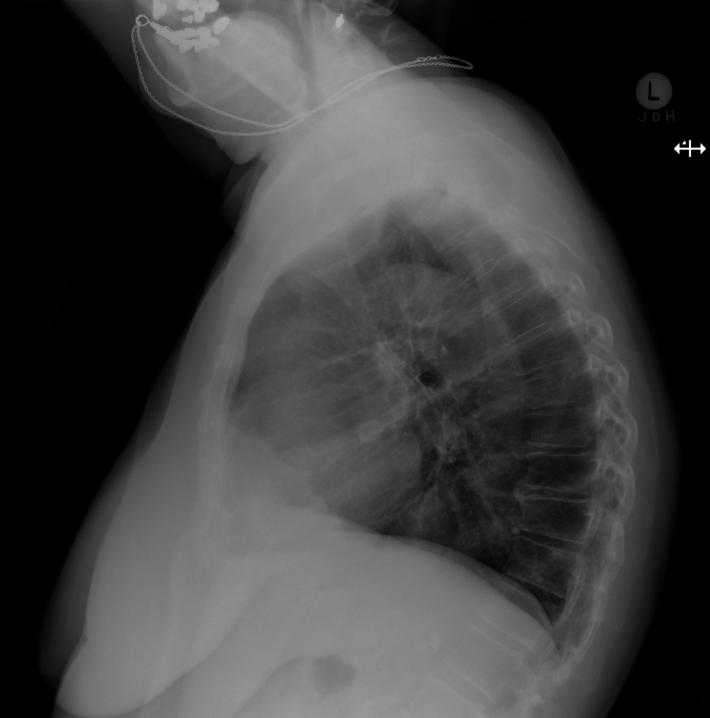}
    \caption{\small An example chest X-ray from the ReXGradient-160K dataset. \textbf{Findings:} The lungs are adequately inflated. There is no focal infiltrate. There is no pleural effusion. No pulmonary parenchymal nodules or masses are observed. The heart and pulmonary vascularity are normal. There is calcification in the wall of the thoracic aorta. The bony thorax exhibits no acute abnormality. \textbf{Impression:} No pulmonary nodules are observed. There is no acute cardiopulmonary abnormality. Thoracic aortic atherosclerosis.}
    \label{fig:rexgradient-example}
  \end{figure}

We embed every finding sentence in the training set with a pretrained sentence transformer, \texttt{all-MiniLM-L6-v2} \cite{reimers2019sentence}, denoted $f_{sent}$. Let $N$ be the total number of unique finding sentences across all training reports. The encoder produces $N$ representations $\tilde{x}_i = f_{sent}(s_i) \in \mathbb{R}^{384}$ for $i=1,\ldots,N$, which we normalize to unit length, $x_i = \tilde{x}_i / \|\tilde{x}_i\|$.

For the report in Figure~\ref{fig:rexgradient-example}, the findings decompose into seven sentences $s_1, \ldots, s_7$: for instance, $s_1$ is ``The lungs are adequately inflated'', $s_2$ is ``There is no focal infiltrate'', and $s_7$ is ``The bony thorax exhibits no acute abnormality''. The corresponding embeddings $x_1, \ldots, x_7$ feed into the next step, whose goal is to group sentences so that each group contains semantically equivalent observations.

\paragraph{Clustering finding sentences:}
With these embeddings in hand, we cluster the sentences so that semantically equivalent observations fall into a common group. We use three algorithms, all with Euclidean distance: K-means, DBSCAN, and HDBSCAN \cite{lloyd1982least, ester1996density, campello2013density}. With default hyperparameters, DBSCAN yields 995 clusters and HDBSCAN yields 3{,}216; for K-means we sweep $k \in \{1000, 2000, 5000\}$.

The two families behave differently. Density-based methods (DBSCAN, HDBSCAN) leave low-density points as \emph{outliers}, so not every sentence is assigned to a cluster. K-means partitions all points and is forced to absorb outliers into the nearest cluster. In our data, K-means assigns all 184{,}535 training sentences to clusters, whereas HDBSCAN clusters only 66{,}477 of them; the remaining sentences are dropped from the enrichment pool. This trade-off matters: density-based methods produce purer clusters at the cost of coverage, while K-means trades purity for coverage. The number of clusters in K-means controls a similar trade-off between expansion reach and cluster noise. To avoid hyperparameter cherry-picking, all clustering methods use library defaults and we report all three values of $k$ for K-means. The choice of sentence encoder is itself a design axis: in Appendix~\ref{sec:appendix-encoder} we hold the pipeline fixed and swap MiniLM for larger general-purpose encoders (Stella, GTE-Qwen2) and biomedical ones (PubMedBERT, BioBERT); every encoder improves over the baseline, and the general-purpose MiniLM remains the strongest on the clinical metrics.


Below we provide examples of sentences that are grouped into the same cluster, demonstrating that semantically similar findings are successfully clustered together:

\begin{tcolorbox}[
  colback=hueA!6,
  colframe=hueA!55,
  boxrule=0.4pt,
  arc=2pt,
  left=4pt, right=4pt, top=3pt, bottom=3pt,
  title={\small\textbf{Cluster 1:} Gray-White Differentiation},
  fonttitle=\small
]
\small\textit{``Gray-white differentiation appears unremarkable.'' / ``The gray-white matter differentiation is within normal limits.'' / ``Gray white differentiation is intact.'' / ...}
\end{tcolorbox}

\noindent Additional cluster examples are provided in Appendix~\ref{sec:appendix-cluster-examples}.

Let $f_C: s_i \mapsto c_i$ denote the sentence-to-cluster-id mapping, where $c_i$ is the cluster id assigned to sentence $s_i$. With $K$ clusters in total, $C = \{c_1, \ldots, c_K\}$, the findings of the $j$-th chest X-ray, which is naturally a set of sentences $\{s_{j1}, s_{j2}, \ldots, s_{jr}\}$ where $r$ is the number of sentences in that report, can therefore be represented as the corresponding set of cluster ids $\{c_{j1}, c_{j2}, \ldots, c_{jr}\} = \{f_C(s_{j1}), \ldots, f_C(s_{jr})\}$. Per-method cluster-size statistics are reported in Appendix~\ref{sec:appendix-hdbscan-stats} an ~\ref{sec:appendix-kmeans-stats}.


\begin{figure*}[ht]
\centering
\resizebox{0.86\textwidth}{!}{%
\begin{tikzpicture}[
    cluster node/.style={circle, draw, minimum size=8mm, font=\footnotesize\bfseries\rmfamily},
    main node/.style={cluster node, fill=blue!25, draw=blue!70, line width=1.5pt},
    bg node1/.style={cluster node, fill=green!20, draw=green!60, line width=1pt},
    bg node2/.style={cluster node, fill=orange!20, draw=orange!60, line width=1pt},
    bg node3/.style={cluster node, fill=gray!15, draw=gray!40},
    main edge/.style={blue!50, line width=1.2pt},
    bg edge1/.style={green!40, line width=0.8pt},
    bg edge2/.style={orange!40, line width=0.8pt},
    bg edge3/.style={gray!30, line width=0.5pt},
    annot/.style={align=left, font=\tiny\rmfamily, fill=white, fill opacity=0.9, text opacity=1, rounded corners=2pt, inner sep=3pt, draw=gray!30},
]

\node[bg node3] (X1) at (-7, 1.5) {$C_{81}$};
\node[bg node3] (X2) at (-6.5, -1.5) {$C_{42}$};
\node[bg node3] (X3) at (7, 1) {$C_{67}$};
\node[bg node3] (X4) at (6.5, -2) {$C_{91}$};
\node[bg node3] (X5) at (-5.5, 3) {$C_{23}$};
\node[bg node3] (X6) at (5.5, 1.5) {$C_{58}$};

\node[bg node1] (G1) at (-6.2, 0) {$C_{12}$};
\node[bg node1] (G2) at (-4.5, 1.5) {$C_{45}$};
\node[bg node1] (G3) at (-4.0, -0.3) {$C_{78}$};
\node[bg node1] (G4) at (-4.5, -1.5) {$C_{33}$};

\draw[bg edge1] (G1) -- (G2);
\draw[bg edge1] (G1) -- (G3);
\draw[bg edge1] (G1) -- (G4);
\draw[bg edge1] (G2) -- (G3);
\draw[bg edge1] (G2) -- (G4);
\draw[bg edge1] (G3) -- (G4);

\node[bg node2] (O1) at (4.2, 0.5) {$C_{19}$};
\node[bg node2] (O2) at (6, 0.5) {$C_{54}$};
\node[bg node2] (O3) at (6.2, -1) {$C_{87}$};

\draw[bg edge2] (O1) -- (O2);
\draw[bg edge2] (O1) -- (O3);
\draw[bg edge2] (O2) -- (O3);

\node[main node] (C1) at (-1.5, 1.7) {$C_1$};
\node[main node] (C2) at (2.8, 1.9) {$C_2$};
\node[main node] (C3) at (2.5, 0) {$C_3$};
\node[main node] (C4) at (-1.5, -0.1) {$C_4$};

\draw[main edge] (C1) -- (C2);
\draw[main edge] (C1) -- (C3);
\draw[main edge] (C1) -- (C4);
\draw[main edge] (C2) -- (C3);
\draw[main edge] (C2) -- (C4);
\draw[main edge] (C3) -- (C4);

\node[annot, anchor=south] at (-1.5, 2.3) {
\textbf{``there is hyperinflation''}\\[1pt]
\textit{``hyperinflation is noted''}\\
\textit{``there is mild hyperinflation''}\\
\textit{``underlying hyperinflation''}
};

\node[annot, anchor=south] at (4.1, 2.4) {
\textbf{``there is mild unchanged cardiomegaly''}\\[1pt]
\textit{``cardiomegaly is noted''}\\
\textit{``cardiomegaly is stable''}\\
\textit{``moderate cardiomegaly''}\\
\textit{``borderline cardiomegaly''}
};

\node[annot, anchor=north] at (3.4, -0.6) {
\textbf{``there are no effusions''}\\
\textit{``no evidence of effusion''}\\
\textit{``no effusion is noted''}\\
\textit{``no sizable effusion is seen''}\\
\textit{``no large effusion is evident''}
};

\node[annot, anchor=north] at (-1.25, -0.8) {
\textbf{``there are upper lobe emphysematous changes''}\\[1pt]
\textit{``possible emphysematous changes in upper lobes''}\\
\textit{``emphysematous changes of the upper lobes''}\\
\textit{``emphysematous change in the right upper lobe''}
};

\node[anchor=north west] at (-7.5, -2.2) {
\begin{tikzpicture}[scale=0.8]
    \node[main node, minimum size=4mm] at (0, 1.35) {};
    \node[anchor=west, font=\footnotesize] at (0.7, 1.1) {Patient 1 findings (4 sentences)};
    
    \node[bg node1, minimum size=4mm] at (0, 0.65) {};
    \node[anchor=west, font=\footnotesize] at (0.7, 0.4) {Patient 2 findings (4 sentences)};
    
    \node[bg node2, minimum size=4mm] at (0, -0.05) {};
    \node[anchor=west, font=\footnotesize] at (0.7, -0.3) {Patient 3 findings (3 sentences)};
    
    \node[bg node3, minimum size=4mm] at (7, 0.7) {};
    \node[anchor=west, font=\footnotesize] at (7.5, 0.5) {Other clusters};
    
    \draw[main edge] (7.0, 1.25) -- (8.0, 1.25);
    \node[anchor=west, font=\footnotesize] at (8.2, 1.2) {Complete subgraph edges};
\end{tikzpicture}
};

\end{tikzpicture}%
}
\caption{\small Visualization of the cluster-based findings representation. 
Each node represents a cluster of semantically similar sentences. 
Patient 1's findings (consisting of multiple sentences) forms a complete subgraph where each node corresponds to the cluster containing one of the finding sentences. 
The blue subgraph shows an example findings with 4 sentences: the bold text shows the original sentence, and italicized text shows other sentences from the same cluster. 
The green and orange subgraphs represent findings from other patients.}
\label{fig:cluster-graph}
\end{figure*}

\subsection{Cluster Polarity Assignment}
We assign each cluster a polarity label: \emph{normal} or \emph{abnormal}. A cluster is normal if its sentences describe healthy, unsuspicious anatomy or non-findings (\textit{e.g.}, ``the lungs are clear''); it is abnormal if its sentences describe a pathological, suspicious, or symptomatic finding (\textit{e.g.}, ``there is a left lower lobe consolidation'').

To label a cluster, we sample one sentence from it at random and prompt a DeepSeek-R1 model with the following instruction:

\begin{tcolorbox}[
  colback=gray!5,
  colframe=gray!50,
  boxrule=0.5pt,
  arc=2pt,
  left=6pt,
  right=6pt,
  top=4pt,
  bottom=4pt,
  fontupper=\small\ttfamily
]
Decide whether the following radiology sentence describes a normal observation or an abnormal observation.\\[4pt]
- If the sentence reports any pathology, suspicious feature, or symptomatic finding, respond with `abnormal'.\\
- If the sentence describes a finding consistent with a healthy patient, respond with `normal'.\\[4pt]
Output only `normal' or `abnormal', with no explanation.\\[4pt]
The sentence is: \{finding\}
\end{tcolorbox}

We repeat this for every cluster, producing a polarity map $h_S: c_i \mapsto s_i$ with $s_i \in \{+1, -1\}$, where $+1$ denotes a normal cluster and $-1$ denotes an abnormal one.

\paragraph{On the noise of single-sentence polarity assignment:}
This single-sentence labelling is noisy and is not a substitute for clinician judgement. It is tolerable here because polarity acts only as a hard \emph{gate} on which clusters may be added (Section~\ref{sec:enrichment}), never as a supervision target: a wrong label only changes whether a cluster is eligible, and mis-labelled clusters are further filtered by the co-occurrence constraint, so errors lower the augmentation rate rather than inject incorrect content. Empirically, even under this noisy gate the gains in Section~\ref{sec:experiments} are large and consistent, indicating that the signal dominates the residual noise; clinician-in-the-loop labelling is a natural refinement. We also directly quantify this noise by labelling member sentences of every inserted normal cluster: the inserted clusters are highly polarity-pure (insertion-weighted purity $\approx99.8\%$), corresponding to an effective abnormal-injection rate of only $\approx0.22\%$ (Appendix~\ref{sec:appendix-purity}).

\subsection{Enrichment via Connected Clusters}\label{sec:enrichment}

We now describe how new sentences are added to a patient's findings. Three definitions formalize the procedure.

\paragraph{Connected Clusters:}
Two clusters $c_i$ and $c_j$ are \emph{connected} if there exists a patient in the training set whose findings contain one sentence from $c_i$ and another from $c_j$. Connectivity is a co-occurrence relation: if two clusters are connected, then the propositions they represent have already been seen together in at least one real report and are not mutually contradictory in that context. In practice we apply a slightly more robust co-occurrence gate governed by two thresholds, fixed across all configurations: two clusters are connected only when they co-occur in at least $\tau_{\text{count}}=3$ reports (a floor on the raw co-occurrence count) and their row-normalized co-occurrence exceeds $\tau_{\text{norm}}=0.001$. This discards spurious single-report co-occurrences; the full procedure is given in Appendix~\ref{sec:maximal_cliques_problem}, Algorithm~\ref{alg:expansion}.

\paragraph{Valid Enrichment:}
Let $F_j = \{c_{j1}, c_{j2}, \ldots, c_{jr}\}$ be the set of clusters spanned by the findings of the $j$-th patient. For a candidate set $F_{cand} \subset C$, we call $(F_j, F_{cand})$ a \emph{valid enrichment} if:

\begin{align}
& \bullet \; F_j \cap F_{cand} = \emptyset, \\
& \bullet \; \forall\, x, y \in F_j \cup F_{cand},\; x, y \text{ are connected}, \\
& \bullet \; \forall\, c \in F_{cand},\; h_S(c) = +1.
\end{align}

Intuitively, a valid enrichment adds only \emph{normal} observations to an existing report, and only when every added cluster co-occurs with every other cluster already present or being added. The motivation is that clinicians focus on abnormalities and frequently omit normal observations they consider uninformative; by definition we never add abnormalities, and connectivity prevents combining mutually contradictory normal findings.

\paragraph{Largest Valid Enrichment:}
$(F_j, F_{cand})$ is a \emph{largest valid enrichment} if it is a valid enrichment and no additional cluster can be added without breaking the conditions above:

\begin{align}
& \bullet \; (F_j, F_{cand}) \text{ is a valid enrichment}, \\
& \bullet \; \forall\, c \in C \setminus (F_j \cup F_{cand}), \nonumber\\
& \quad (F_j \cup F_{cand}, \{c\}) \text{ is not a valid enrichment}.
\end{align}

In other words, we add as many normal sentences as possible without violating connectivity or polarity. Figure~\ref{fig:enrichment-graph} illustrates a small example. Finding possible valid enrichments is very closely related to classical \emph{maximal clique enumeration} problem in graph theory. The algorithmic details of finding the largest valid enrichment are deferred to Appendix~\ref{sec:maximal_cliques_problem}.

Like the polarity assignment, this enrichment is inherently noisy; however, the large and consistent improvements in Section~\ref{sec:experiments} experimentally show that the signal it provides far outweighs that noise.


\begin{figure*}[ht]
\centering
\resizebox{0.86\textwidth}{!}{%
\begin{tikzpicture}[
    base node/.style={circle, draw, minimum size=12mm, font=\small\bfseries\rmfamily, line width=1.2pt},
    normal node/.style={base node, fill=green!20, draw=green!60},
    abnormal node/.style={base node, fill=red!20, draw=red!60},
    candidate normal/.style={base node, fill=green!15, draw=green!50, dashed},
    candidate abnormal/.style={base node, fill=red!15, draw=red!50, dashed},
    candidate invalid/.style={base node, fill=gray!15, draw=gray!50, dashed},
    connected/.style={black, line width=1pt},
    candidate edge/.style={green!60, line width=0.8pt, densely dotted},
    invalid edge/.style={red!40, line width=0.8pt, densely dashed},
    annot/.style={font=\scriptsize\rmfamily, align=center},
    main box/.style={rounded corners=8pt, draw=blue!50, line width=1.5pt, fill=blue!5, inner sep=8pt},
]

\node[font=\bfseries] at (-4.5, 4.5) {Base Findings + Candidates};

\draw[main box] (-6.3, -0.8) rectangle (-2.7, 2.8);
\node[font=\scriptsize\itshape, blue!60] at (-4.5, 3.1) {Base Findings $F_j$};

\node[normal node] (C1) at (-5.5, 2) {$c_1$};
\node[abnormal node] (C2) at (-3.5, 2) {$c_2$};
\node[abnormal node] (C3) at (-5.5, 0) {$c_3$};
\node[normal node] (C4) at (-3.5, 0) {$c_4$};

\draw[connected] (C1) -- (C2);
\draw[connected] (C1) -- (C3);
\draw[connected] (C1) -- (C4);
\draw[connected] (C2) -- (C3);
\draw[connected] (C2) -- (C4);
\draw[connected] (C3) -- (C4);

\node[candidate normal] (C5) at (-7.5, 1) {$c_5$};
\node[candidate abnormal] (C6) at (-1.5, 2.5) {$c_6$};
\node[candidate invalid] (C7) at (-1.5, -0.5) {$c_7$};
\node[candidate normal] (C8) at (-5.5, -2.5) {$c_8$};
\node[candidate normal] (C9) at (-3.5, -2.5) {$c_9$};

\draw[candidate edge] (C5) -- (C1);
\draw[candidate edge] (C5) -- (C2);
\draw[candidate edge] (C5) -- (C3);
\draw[candidate edge] (C5) -- (C4);

\draw[invalid edge] (C6) -- (C1);
\draw[invalid edge] (C6) -- (C2);
\draw[invalid edge] (C6) -- (C4);

\draw[candidate edge] (C7) -- (C2);
\draw[candidate edge] (C7) -- (C4);

\draw[candidate edge] (C8) -- (C1);
\draw[candidate edge] (C8) -- (C2);
\draw[candidate edge] (C8) -- (C3);
\draw[candidate edge] (C8) -- (C4);
\draw[candidate edge] (C9) -- (C1);
\draw[candidate edge] (C9) -- (C2);
\draw[candidate edge] (C9) -- (C3);
\draw[candidate edge] (C9) -- (C4);
\draw[candidate edge, line width=1pt] (C8) -- (C9);


\node[annot, anchor=east] at (-8.2, 1) {\textcolor{green!60!black}{\checkmark} Valid};
\node[annot, anchor=west] at (-0.8, 2.5) {\textcolor{red!60!black}{$\times$} Abnormal};
\node[annot, anchor=west] at (-0.8, -0.5) {\textcolor{red!60!black}{$\times$} Not connected};
\node[annot, anchor=north] at (-4.5, -3.2) {\textcolor{green!60!black}{\checkmark} Valid pair};

\node[font=\bfseries] at (5, 4.5) {Valid Enrichments};

\node[font=\small\bfseries] at (2.5, 3.5) {Enrichment 1:};

\node[normal node, minimum size=8mm] (E1C1) at (1.5, 2) {$c_1$};
\node[abnormal node, minimum size=8mm] (E1C2) at (2.5, 2.7) {$c_2$};
\node[abnormal node, minimum size=8mm] (E1C3) at (2.5, 1.3) {$c_3$};
\node[normal node, minimum size=8mm] (E1C4) at (3.5, 2) {$c_4$};
\node[normal node, minimum size=8mm, fill=green!40, draw=green!70, line width=1.5pt] (E1C5) at (2.5, 0.3) {$c_5$};

\draw[connected, line width=0.8pt] (E1C1) -- (E1C2);
\draw[connected, line width=0.8pt] (E1C1) -- (E1C3);
\draw[connected, line width=0.8pt] (E1C1) -- (E1C4);
\draw[connected, line width=0.8pt] (E1C2) -- (E1C3);
\draw[connected, line width=0.8pt] (E1C2) -- (E1C4);
\draw[connected, line width=0.8pt] (E1C3) -- (E1C4);
\draw[green!60, line width=1pt] (E1C5) -- (E1C1);
\draw[green!60, line width=1pt] (E1C5) -- (E1C2);
\draw[green!60, line width=1pt] (E1C5) -- (E1C3);
\draw[green!60, line width=1pt] (E1C5) -- (E1C4);

\node[annot] at (2.5, -0.5) {$F_j \cup \{c_5\}$};

\node[font=\small\bfseries] at (6.5, 3.5) {Enrichment 2:};

\node[normal node, minimum size=8mm] (E2C1) at (5.5, 2) {$c_1$};
\node[abnormal node, minimum size=8mm] (E2C2) at (6.5, 2.7) {$c_2$};
\node[abnormal node, minimum size=8mm] (E2C3) at (6.5, 1.3) {$c_3$};
\node[normal node, minimum size=8mm] (E2C4) at (7.5, 2) {$c_4$};
\node[normal node, minimum size=8mm, fill=green!40, draw=green!70, line width=1.5pt] (E2C8) at (5.8, 0.3) {$c_8$};
\node[normal node, minimum size=8mm, fill=green!40, draw=green!70, line width=1.5pt] (E2C9) at (7.2, 0.3) {$c_9$};

\draw[connected, line width=0.8pt] (E2C1) -- (E2C2);
\draw[connected, line width=0.8pt] (E2C1) -- (E2C3);
\draw[connected, line width=0.8pt] (E2C1) -- (E2C4);
\draw[connected, line width=0.8pt] (E2C2) -- (E2C3);
\draw[connected, line width=0.8pt] (E2C2) -- (E2C4);
\draw[connected, line width=0.8pt] (E2C3) -- (E2C4);
\draw[green!60, line width=1pt] (E2C8) -- (E2C1);
\draw[green!60, line width=1pt] (E2C8) -- (E2C2);
\draw[green!60, line width=1pt] (E2C8) -- (E2C3);
\draw[green!60, line width=1pt] (E2C8) -- (E2C4);
\draw[green!60, line width=1pt] (E2C9) -- (E2C1);
\draw[green!60, line width=1pt] (E2C9) -- (E2C2);
\draw[green!60, line width=1pt] (E2C9) -- (E2C3);
\draw[green!60, line width=1pt] (E2C9) -- (E2C4);
\draw[green!70, line width=1.2pt] (E2C8) -- (E2C9);

\node[annot] at (6.5, -0.5) {$F_j \cup \{c_8, c_9\}$};

\node[font=\small\bfseries] at (5, -1.5) {Invalid Candidates:};

\node[candidate abnormal, minimum size=8mm] at (2.5, -2.5) {$c_6$};
\node[annot, anchor=west] at (3.2, -2.5) {$h_S(c_6) = -1$ (abnormal)};

\node[candidate invalid, minimum size=8mm] at (2.5, -3.8) {$c_7$};
\node[annot, anchor=west] at (3.2, -3.8) {Not connected to $c_1, c_3$};

\node[anchor=north west] at (-7.5, -4.2) {
\begin{tikzpicture}[scale=0.75]
    \node[normal node, minimum size=4mm] at (0, 0.2) {};
    \node[anchor=west, font=\scriptsize] at (0.5, 0) {Normal cluster};

    \node[abnormal node, minimum size=4mm] at (3.5, 0.2) {};
    \node[anchor=west, font=\scriptsize] at (4, 0) {Abnormal cluster};
    
    \draw[connected] (7, 0) -- (8, 0);
    \node[anchor=west, font=\scriptsize] at (8.2, 0) {Connected};
    
    \draw[candidate edge] (0, -0.7) -- (1, -0.7);
    \node[anchor=west, font=\scriptsize] at (1.2, -0.7) {Candidate edge};
    
    \draw[main box, inner sep=2pt] (5, -0.9) rectangle (6, -0.5);
    \node[anchor=west, font=\scriptsize] at (6.2, -0.7) {Base findings};
\end{tikzpicture}
};

\end{tikzpicture}%
}
\caption{\small Illustration of the valid enrichment process. \textbf{Left:} the base findings $F_j = \{c_1, c_2, c_3, c_4\}$ (inside the blue box) and candidate clusters (dashed nodes). Green nodes are normal clusters; red nodes are abnormal clusters. Candidate $c_5$ is valid (normal and connected to all base clusters). Candidate $c_6$ is invalid (abnormal). Candidate $c_7$ is invalid (not connected to $c_1$ and $c_3$). Candidates $c_8$ and $c_9$ form a valid pair (both normal, connected to all base clusters and to each other). Note that $c_5$ is not connected to $\{c_8, c_9\}$, so they cannot belong to the same enrichment. \textbf{Right:} two largest valid enrichments, $F_j \cup \{c_5\}$ and $F_j \cup \{c_8, c_9\}$. $F_j \cup \{c_8\}$ is a valid enrichment but not a \emph{largest} one.}
\label{fig:enrichment-graph}
\end{figure*}

\paragraph{Enrichment examples:}
We now show a real example of a valid enrichment from our training data, pairing an original set of findings with the expansion clusters that the procedure adds. The inserted sentences are semantically consistent with the original report, and the procedure never introduces an abnormal finding.

\begin{tcolorbox}[
  colback=hueF!7,
  colframe=hueF!60,
  boxrule=0.5pt,
  arc=2pt,
  title={\textbf{Enrichment Example (Multiple Expansions)}},
  fonttitle=\small\bfseries
]
\textbf{Initial Findings:}
\begin{itemize}[noitemsep,topsep=2pt,leftmargin=*]
  \item Right hilar mass is suspected.
  \item Heart is normal in size. Lungs are clear.
  \item No pneumothorax or effusion.
  \item Airspace disease is present in the inferior right upper lobe.
\end{itemize}
\textbf{Expansion Cluster 1:} \textit{``visualized bony thorax intact'', ``the visualized bony structures of the thorax are intact'', ...}\\
\textbf{Expansion Cluster 2:} \textit{``normal lung volumes'', ``lung volumes are normal'', ``the lung volumes are normal'', ...}
\end{tcolorbox}

\noindent The same initial findings can yield several distinct valid enrichments depending on which connected normal clusters are sampled; further alternatives and examples are given in Appendix~\ref{sec:appendix-enrichment-examples}. Appendix~\ref{sec:appendix-recovered-normals} additionally reports which normal findings the procedure most often recovers -- canonical omissions such as \emph{normal heart size} and \emph{no pneumothorax} -- together with the fraction of reports that are enriched.

\subsection{Training: SFT and Cluster-Aware GRPO}

\begin{figure}[t]
\centering
\input{figs/paper_style}
\resizebox{\columnwidth}{!}{%
\begin{tikzpicture}[font=\sffamily]


  \node[imgframe] (xray) at (0,2.55)
       {\includegraphics[width=1.15cm]{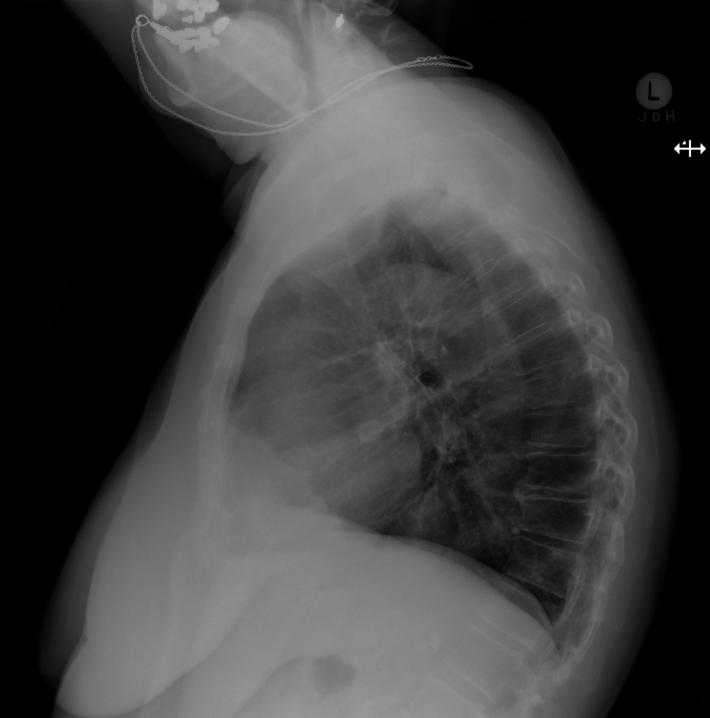}};
  \node[font=\scriptsize, color=inkmed!85] at (0,1.78) {Chest X-ray $v$};

  \begin{scope}[shift={(1.55,1.75)}]
    \trapezoidNarr{hueB}{ViT}{$224^2$}{$d_v$}
  \end{scope}
  \node[font=\tiny, color=hueB!55!black] at (2.15,1.55) {vision encoder};
  \draw[arw] (xray.east) -- (1.60,2.55);

  \begin{scope}[shift={(3.55,1.75)}]
    \trapezoidExp{hueC}{Proj}{$d_v$}{$d$}
  \end{scope}
  \node[font=\tiny, color=hueC!50!black] at (4.10,1.55) {linear};
  \draw[arw] (2.75,2.55) -- (3.62,2.55);

  \tokenrow{5.05}{1.95}{4}{hueE}{Image tokens}{$I_{1:m}$}
  \draw[arw] (4.80,2.55) -- (4.96,2.55);

  \node[draw=hueA!70, fill=hueA!10, rounded corners=3pt, line width=0.6pt,
        align=center, text=hueA!45!black, font=\scriptsize,
        minimum width=1.7cm, minimum height=0.95cm] (txt) at (3.95,0.20)
        {\textbf{Findings}\\(text prompt)};
  \tokenrow{5.05}{-0.60}{6}{hueA}{Text tokens}{$y_{1:n}$}
  \draw[arw] (txt.east) -- (4.96,0.20);

  \node[sumnode] (cat) at (7.05,1.30) {$+$};
  \draw[thinarw] (5.73,2.55) -- (7.05,2.55) -- (cat.north);
  \draw[thinarw] (6.05,0.15) -- (7.05,0.15) -- (cat.south);

  \begin{scope}[shift={(8.05,0.55)}]
    \layerstack{6}{hueD}{1.5}
  \end{scope}
  \node[font=\small\bfseries, color=hueD!55!black] at (8.80,2.55) {LLM};
  \node[font=\tiny, color=hueD!50!black] at (8.80,0.28) {LoRA-tuned};
  \draw[arw] (cat.east) -- (8.00,1.30);

  \tokenrow{10.85}{0.55}{5}{hueF}{Predicted findings}{$\hat{y}_{1:n}$}
  \draw[arw] (9.60,1.30) -- (10.76,1.30);

  \node[font=\normalsize, color=emph] (loss) at (13.15,1.30)
        {$\mathcal{L}_{\mathrm{NTP}}$};
  \draw[lossarw] (11.95,1.30) -- (loss.west);

\end{tikzpicture}%
}
\caption{\small Model architecture. The chest X-ray is encoded by a
Vision Transformer and mapped by a linear projector into $m$ image tokens
$I_{1:m}$. These are concatenated with the tokenized findings $y_{1:n}$ and
passed to a LoRA-tuned large language model, trained with the next-token
prediction loss on the finding tokens.}
\label{fig:model-architecture}
\end{figure}

\paragraph{Supervised fine-tuning:}
The backbone follows the standard vision--language recipe (Figure~\ref{fig:model-architecture}): a Vision Transformer encodes the chest X-ray and a linear layer projects its features into $m$ image tokens $I_{1:m}$, which are concatenated with the tokenized findings $y_{1:n}$. We train with the standard next-token-prediction objective, masking the image tokens in the loss so the model is only penalized for predicting finding tokens conditioned on the image representation:
\begin{equation}
\mathcal{L} = -\frac{1}{n}\sum_{t=1}^{n} \log P_\theta(y_t \mid I_{1:m},\, y_{1:t-1}),
\label{eq:ntp-loss}
\end{equation}
where $\theta$ are the trainable parameters and $P_\theta(\cdot)$ is the LLM's next-token distribution.


\paragraph{Cluster-aware GRPO reward:}
\label{sec:grpo_rewarding}
Our clustering can also be reused as a reward signal in RL post-training, independently of enrichment. We adopt the same framing as GRPO over reasoning models: the input is formatted as {\footnotesize\ttfamily <think>findings</think>\allowbreak <answer>impression</answer>}, and the standard reward is the exact match between the generated and ground-truth impressions, $\mathcal{R}_{exact} = \mathbf{1}[y_{gt} = y_{gen}]$.

We add a complementary reward based on the cluster assignments of the generated findings. Let $F_{gt} = \{c_1, \ldots, c_l\}$ denote the set of clusters spanned by the ground-truth findings of a sample, and $F_{gen} = \{c_{g1}, \ldots, c_{gm}\}$ the set spanned by the generated findings. The cluster-aware reward is the F1 between these two sets, added to the exact-match reward:
\begin{equation}
\mathcal{R}_{cl} = \frac{2\, |F_{gt} \cap F_{gen}|}{|F_{gt}| + |F_{gen}|} + \mathcal{R}_{exact}.
\end{equation}

In experiments, we compare $\mathcal{R}_{cl}$ to $\mathcal{R}_{exact}$ alone. This is orthogonal to the enrichment in Section~\ref{sec:enrichment}: enrichment changes the training data, while the cluster-aware reward changes the supervision signal in RL.

\section{Experiments}\label{sec:experiments}

We evaluate SemEnrich on the ReXGradient-160K dataset \cite{zhang2025rexgradient}. We use the official patient-disjoint splits, with train/validation/test $=95{,}716\,/\,6{,}964\,/\,6{,}807$ patients and zero patient overlap between splits. Clustering, cluster polarities, and the co-occurrence connectivity graph are all built from the \emph{training} split only. For each configuration, a baseline model is trained on the raw dataset, and an enriched model is trained on the same dataset with our augmentation applied at batch time: for every sample, we draw uniformly at random one of its precomputed largest valid enrichments and append the corresponding sentences to the findings before the forward pass. The evaluation splits are never enriched and the test set is never modified, so there is no train--test leakage and all the performance improvements experimentally prove a richer self-supervision.

\paragraph{Backbones:}
We use ViT-B/16 as the vision encoder with input resolution $224\!\times\!224$. We evaluate five LLM backbones: DeepSeek-R1-Distill-Qwen-1.5B and DeepSeek-R1-Distill-Llama-8B \cite{guo2025deepseek}, Qwen3-8B \cite{yang2025qwen3}, Mistral-7B-v0.1 \cite{jiang2023mistral7b}, and Gemma3-4B \cite{gemmateam2025gemma3}.

\paragraph{Training:}
We fine-tune with LoRA (rank $r{=}8$) for 25 epochs using AdamW with learning rate $10^{-4}$ and batch size 144. All experiments run on a single NVIDIA A100 GPU.

\begin{table*}[t]
  \centering
  \caption{Main results across five LLM backbones (mean$\pm$std over five seeds). Each block is one metric (in percent); the best method per backbone is bolded and the method with the best mean is shaded. Additional results are in Appendix~\ref{sec:appendix-additional_results}.}
  \label{tab:main_results}
  \footnotesize
  \setlength{\tabcolsep}{4pt}
  \begin{tabular}{llcccccc}
    \toprule
    Metric & Method & DS-R1-Qwen-1.5B & Gemma3-4B & DS-R1-Llama-8B & Qwen3-8B & Mistral-7B & \textit{Mean} \\
    \midrule
    \multirow{6}{1.7cm}{\centering Sentence-BLEU-1} & Baseline & 27.58$\pm$0.87 & 21.63$\pm$0.70 & 27.45$\pm$0.86 & 28.11$\pm$1.17 & 29.87$\pm$1.67 & 26.93 \\
     & K-means-1000 & 26.40$\pm$0.66 & 22.80$\pm$0.74 & 26.85$\pm$1.00 & 26.88$\pm$0.71 & 23.12$\pm$0.52 & 25.21 \\
     & K-means-2000 & 28.74$\pm$0.97 & 22.25$\pm$0.70 & 29.09$\pm$0.77 & 29.36$\pm$0.94 & 29.58$\pm$0.69 & 27.80 \\
     & K-means-5000 & 28.16$\pm$0.62 & 23.32$\pm$0.32 & \textbf{29.38$\pm$0.61} & 29.40$\pm$0.79 & 32.53$\pm$1.15 & 28.56 \\
     & DBSCAN & 27.03$\pm$1.31 & 21.36$\pm$0.83 & 28.44$\pm$0.69 & 28.02$\pm$0.56 & 28.89$\pm$1.00 & 26.75 \\
     & \cellcolor{gray!18}HDBSCAN & \cellcolor{gray!18}\textbf{29.51$\pm$0.81} & \cellcolor{gray!18}\textbf{23.83$\pm$0.91} & \cellcolor{gray!18}29.11$\pm$0.54 & \cellcolor{gray!18}\textbf{30.35$\pm$1.35} & \cellcolor{gray!18}\textbf{32.64$\pm$1.00} & \cellcolor{gray!18}\textbf{29.09} \\
    \midrule
    \multirow{6}{1.7cm}{\centering COMET} & Baseline & 63.68$\pm$0.49 & 62.05$\pm$0.53 & 63.95$\pm$0.48 & 63.80$\pm$0.80 & 63.59$\pm$0.77 & 63.41 \\
     & K-means-1000 & 65.76$\pm$0.45 & 63.65$\pm$0.39 & 66.66$\pm$0.37 & 66.32$\pm$0.40 & 65.42$\pm$0.26 & 65.56 \\
     & K-means-2000 & 67.14$\pm$0.43 & 63.51$\pm$0.32 & 67.37$\pm$0.37 & 67.40$\pm$0.49 & 68.10$\pm$0.40 & 66.70 \\
     & K-means-5000 & 66.38$\pm$0.49 & 63.97$\pm$0.22 & 67.02$\pm$0.36 & 67.16$\pm$0.52 & 67.33$\pm$0.51 & 66.37 \\
     & DBSCAN & 63.28$\pm$0.65 & 61.93$\pm$0.35 & 64.24$\pm$0.59 & 64.09$\pm$0.63 & 63.65$\pm$0.60 & 63.44 \\
     & \cellcolor{gray!18}HDBSCAN & \cellcolor{gray!18}\textbf{67.34$\pm$0.41} & \cellcolor{gray!18}\textbf{64.59$\pm$0.65} & \cellcolor{gray!18}\textbf{67.44$\pm$0.31} & \cellcolor{gray!18}\textbf{67.71$\pm$0.73} & \cellcolor{gray!18}\textbf{69.16$\pm$0.57} & \cellcolor{gray!18}\textbf{67.25} \\
    \midrule
    \multirow{6}{1.7cm}{\centering BERTScore-F1} & Baseline & 67.60$\pm$0.48 & 65.35$\pm$0.31 & 67.83$\pm$0.44 & 67.79$\pm$0.67 & 67.86$\pm$0.75 & 67.29 \\
     & K-means-1000 & 68.62$\pm$0.37 & \textbf{66.93$\pm$0.41} & 69.20$\pm$0.51 & 69.11$\pm$0.34 & 67.92$\pm$0.28 & 68.35 \\
     & K-means-2000 & 69.54$\pm$0.66 & 66.38$\pm$0.24 & 69.92$\pm$0.50 & 69.85$\pm$0.44 & 69.14$\pm$0.58 & 68.97 \\
     & K-means-5000 & 69.15$\pm$0.47 & 66.72$\pm$0.13 & \textbf{69.97$\pm$0.39} & 69.87$\pm$0.36 & \textbf{71.40$\pm$0.59} & 69.42 \\
     & DBSCAN & 66.94$\pm$0.66 & 65.17$\pm$0.31 & 67.74$\pm$0.53 & 67.54$\pm$0.34 & 67.09$\pm$0.65 & 66.89 \\
     & \cellcolor{gray!18}HDBSCAN & \cellcolor{gray!18}\textbf{69.66$\pm$0.42} & \cellcolor{gray!18}66.80$\pm$0.60 & \cellcolor{gray!18}69.80$\pm$0.34 & \cellcolor{gray!18}\textbf{70.07$\pm$0.69} & \cellcolor{gray!18}71.35$\pm$0.56 & \cellcolor{gray!18}\textbf{69.54} \\
    \midrule
    \multirow{6}{1.7cm}{\centering CheXbert-14} & Baseline & 36.38$\pm$1.97 & 27.12$\pm$1.93 & \textbf{41.34$\pm$2.37} & 37.98$\pm$3.77 & 37.34$\pm$2.22 & 36.03 \\
     & K-means-1000 & 35.59$\pm$2.12 & 25.11$\pm$1.54 & 39.60$\pm$2.78 & 37.23$\pm$2.21 & 21.55$\pm$1.54 & 31.82 \\
     & K-means-2000 & 35.90$\pm$2.98 & 19.93$\pm$1.23 & 38.78$\pm$1.24 & 38.17$\pm$2.77 & 38.43$\pm$2.70 & 34.24 \\
     & K-means-5000 & 35.17$\pm$2.99 & 25.11$\pm$1.66 & 40.18$\pm$2.51 & 38.80$\pm$3.42 & \textbf{39.30$\pm$3.31} & 35.71 \\
     & DBSCAN & 34.94$\pm$2.01 & 22.33$\pm$1.21 & 38.32$\pm$2.03 & 37.74$\pm$2.77 & 38.23$\pm$1.96 & 34.31 \\
     & \cellcolor{gray!18}HDBSCAN & \cellcolor{gray!18}\textbf{36.52$\pm$1.69} & \cellcolor{gray!18}\textbf{29.69$\pm$3.32} & \cellcolor{gray!18}39.72$\pm$2.38 & \cellcolor{gray!18}\textbf{39.31$\pm$1.44} & \cellcolor{gray!18}39.25$\pm$2.71 & \cellcolor{gray!18}\textbf{36.90} \\
    \midrule
    \multirow{6}{1.7cm}{\centering RadGraph-F1} & Baseline & 26.51$\pm$1.00 & 18.49$\pm$1.07 & 26.08$\pm$1.18 & 27.32$\pm$1.69 & 30.21$\pm$1.72 & 25.72 \\
     & K-means-1000 & 24.93$\pm$1.03 & 20.62$\pm$0.82 & 26.07$\pm$1.40 & 25.97$\pm$1.35 & 23.69$\pm$1.02 & 24.26 \\
     & K-means-2000 & \textbf{28.25$\pm$1.62} & 18.90$\pm$0.93 & \textbf{28.34$\pm$1.28} & 28.69$\pm$1.47 & 31.45$\pm$1.47 & 27.13 \\
     & K-means-5000 & 26.04$\pm$1.00 & 19.83$\pm$0.97 & 27.92$\pm$0.99 & 28.32$\pm$1.22 & 31.65$\pm$1.72 & 26.75 \\
     & DBSCAN & 24.03$\pm$1.60 & 16.23$\pm$1.22 & 25.16$\pm$1.04 & 24.78$\pm$1.14 & 28.73$\pm$1.61 & 23.79 \\
     & \cellcolor{gray!18}HDBSCAN & \cellcolor{gray!18}27.90$\pm$1.22 & \cellcolor{gray!18}\textbf{20.71$\pm$1.55} & \cellcolor{gray!18}28.07$\pm$0.80 & \cellcolor{gray!18}\textbf{29.28$\pm$2.00} & \cellcolor{gray!18}\textbf{32.23$\pm$1.60} & \cellcolor{gray!18}\textbf{27.64} \\
    \bottomrule
  \end{tabular}
\end{table*}

\subsection{Main Results}

Table~\ref{tab:main_results} compares baseline training with five clustering variants (DBSCAN, HDBSCAN, and K-means with $k{\in}\{1000, 2000, 5000\}$) across six metrics. We report Sentence-BLEU-1 for lexical overlap; COMET, which penalizes irrelevant or repetitive content and is therefore the metric most sensitive to noisy enrichment in our setting; BERTScore-F1 for contextual semantic similarity; and the clinically grounded CheXbert-F1 (disease-label agreement) and RadGraph-F1 (entity--relation-graph agreement). The pattern is consistent across all of them: HDBSCAN is the strongest clustering choice and has the best mean on every metric and K-means-5000 is a close second. Corpus-BLEU, METEOR, ChrF++, and ROUGE are deferred to Appendix~\ref{sec:appendix-additional_results} and show the same qualitative trend; $n$-gram corpus-level scores favor longer outputs and, since our method adds sentences, should be read with caution, so we rely primarily on the semantic and clinical metrics here.

Although this work focuses on improving the training data rather than on producing the single best model, for completeness we situate our absolute numbers against open-source state-of-the-art medical vision--language models in Appendix~\ref{sec:appendix-sota}. These comparisons only locate where a report-only fine-tune lands relative to released checkpoints; the evidence for the enrichment itself comes from the controlled baseline-vs-enriched comparison (Table~\ref{tab:main_results}) and the size-matched random-expansion ablation (Section~\ref{sec:ablation}), not from the SOTA table.

\paragraph{Generalization to a second dataset:}
To test whether the enrichment transfers beyond ReXGradient-160K, we additionally run the full SemEnrich pipeline on MIMIC-CXR \cite{johnson2019mimic}. Despite MIMIC's single-institution origin and noisier reports, the pipeline transfers cleanly and HDBSCAN enrichment again improves a Mistral-7B backbone across lexical, semantic, and clinical metrics; the full setup and numbers are in Appendix~\ref{sec:appendix-mimic}, and a comparison against medical VLMs on MIMIC is in Appendix~\ref{sec:appendix-mimic-sota}.

\subsection{Ablation Study}
\label{sec:ablation}

To isolate the contribution of semantic clustering from that of simply adding more sentences, we compare three settings: the raw baseline, HDBSCAN-based enrichment (our method), and a \emph{random-expansion} ablation in which we keep the same expansion size as our method but replace each added sentence with one drawn uniformly at random from a different cluster, ignoring connectivity and polarity.

As shown in Table~\ref{tab:ablation_results}, HDBSCAN enrichment improves over the baseline on every backbone and every metric, while random expansion is consistently \emph{below} the baseline. The gain therefore stems from the semantic structure of the clusters, not from the additional sentence count alone.

\begin{table*}[t]
  \centering
  \caption{Ablation: semantic (HDBSCAN) enrichment vs.\ random expansion (mean$\pm$std over five seeds, percent). The best method per backbone is bolded and the method with the best mean is shaded. Sentence-BLEU-1 and CheXbert-5 results are in Appendix~\ref{sec:appendix-additional_results}.}
  \label{tab:ablation_results}
  \footnotesize
  \setlength{\tabcolsep}{4pt}
  \begin{tabular}{llcccccc}
    \toprule
    Metric & Method & DS-R1-Qwen-1.5B & Gemma3-4B & DS-R1-Llama-8B & Qwen3-8B & Mistral-7B & \textit{Mean} \\
    \midrule
    \multirow{3}{1.7cm}{\centering COMET} & Baseline & 63.68$\pm$0.49 & 62.05$\pm$0.53 & 63.95$\pm$0.48 & 63.80$\pm$0.80 & 63.59$\pm$0.77 & 63.41 \\
     & Random expansion & 63.97$\pm$0.48 & 61.97$\pm$0.41 & 64.53$\pm$0.57 & 64.35$\pm$0.28 & 61.53$\pm$0.47 & 63.27 \\
     & \cellcolor{gray!18}HDBSCAN & \cellcolor{gray!18}\textbf{67.34$\pm$0.41} & \cellcolor{gray!18}\textbf{64.59$\pm$0.65} & \cellcolor{gray!18}\textbf{67.44$\pm$0.31} & \cellcolor{gray!18}\textbf{67.71$\pm$0.73} & \cellcolor{gray!18}\textbf{69.16$\pm$0.57} & \cellcolor{gray!18}\textbf{67.25} \\
    \midrule
    \multirow{3}{1.7cm}{\centering BERTScore-F1} & Baseline & 67.60$\pm$0.48 & 65.35$\pm$0.31 & 67.83$\pm$0.44 & 67.79$\pm$0.67 & 67.86$\pm$0.75 & 67.29 \\
     & Random expansion & 67.06$\pm$0.55 & 65.43$\pm$0.31 & 67.91$\pm$0.55 & 67.62$\pm$0.11 & 66.50$\pm$0.57 & 66.91 \\
     & \cellcolor{gray!18}HDBSCAN & \cellcolor{gray!18}\textbf{69.66$\pm$0.42} & \cellcolor{gray!18}\textbf{66.80$\pm$0.60} & \cellcolor{gray!18}\textbf{69.80$\pm$0.34} & \cellcolor{gray!18}\textbf{70.07$\pm$0.69} & \cellcolor{gray!18}\textbf{71.35$\pm$0.56} & \cellcolor{gray!18}\textbf{69.54} \\
    \midrule
    \multirow{3}{1.7cm}{\centering RadGraph-F1} & Baseline & 26.51$\pm$1.00 & 18.49$\pm$1.07 & 26.08$\pm$1.18 & 27.32$\pm$1.69 & 30.21$\pm$1.72 & 25.72 \\
     & Random expansion & 23.57$\pm$1.14 & 17.90$\pm$1.02 & 24.79$\pm$1.50 & 24.28$\pm$0.74 & 23.99$\pm$1.20 & 22.90 \\
     & \cellcolor{gray!18}HDBSCAN & \cellcolor{gray!18}\textbf{27.90$\pm$1.22} & \cellcolor{gray!18}\textbf{20.71$\pm$1.55} & \cellcolor{gray!18}\textbf{28.07$\pm$0.80} & \cellcolor{gray!18}\textbf{29.28$\pm$2.00} & \cellcolor{gray!18}\textbf{32.23$\pm$1.60} & \cellcolor{gray!18}\textbf{27.64} \\
    \midrule
    \multirow{3}{1.7cm}{\centering CheXbert-14} & Baseline & 36.38$\pm$1.97 & 27.12$\pm$1.93 & \textbf{41.34$\pm$2.37} & 37.98$\pm$3.77 & 37.34$\pm$2.22 & 36.03 \\
     & Random expansion & 30.04$\pm$1.32 & 23.12$\pm$2.55 & 28.85$\pm$2.38 & 33.63$\pm$3.77 & 25.15$\pm$1.38 & 28.16 \\
     & \cellcolor{gray!18}HDBSCAN & \cellcolor{gray!18}\textbf{36.52$\pm$1.69} & \cellcolor{gray!18}\textbf{29.69$\pm$3.32} & \cellcolor{gray!18}39.72$\pm$2.38 & \cellcolor{gray!18}\textbf{39.31$\pm$1.44} & \cellcolor{gray!18}\textbf{39.25$\pm$2.71} & \cellcolor{gray!18}\textbf{36.90} \\
    \bottomrule
  \end{tabular}
\end{table*}

\subsection{GRPO Rewarding Results}
We now evaluate the cluster-aware reward $\mathcal{R}_{cl}$ introduced in Section \ref{sec:grpo_rewarding} against the standard exact-match reward $\mathcal{R}_{exact}$, in a GRPO stage that follows SFT. Recall that $\mathcal{R}_{cl}$ adds the F1 score between the ground-truth and generated cluster sets for the \emph{findings} section on top of the exact-match reward for the \emph{impression} section; we use HDBSCAN clustering throughout.

The results in Table \ref{tab:grpo_combined} are computed on the impression section only. This is a deliberately conservative protocol: our reward augments the findings part, and we test whether the gain transfers to the impression -- the final conclusion of the report -- which would indicate that the model has produced more informative findings as a by-product. Each run is averaged over 5 random seeds. Across both backbones and all three metrics, $\mathcal{R}_{cl}$ exceeds $\mathcal{R}_{exact}$ (+$2.78\%$ COMET, +$3.14\%$ BERTScore, +$14.21\%$ ROUGE-L on average). Additional GRPO details are in Appendix \ref{sec:grpo_training_additional_results}.

\begin{table}[H]
    \caption{GRPO results comparing cluster-based reward $\mathcal{R}_{cl}$ and exact match reward $\mathcal{R}_{exact}$ on the impression section. Additional results are in Appendix~\ref{sec:grpo_training_additional_results}.}
    \label{tab:grpo_combined}
    \centering
    \resizebox{\columnwidth}{!}{%
    \begin{tabular}{l l cc}
        \toprule
        Metric & Model & $\mathcal{R}_{cl}$ & $\mathcal{R}_{exact}$ \\
        \midrule
        \multirow{2}{*}{COMET}
          & Qwen2.5-VL-3B & \textbf{0.6179$\pm$0.003} & 0.5973$\pm$0.002 \\
          & Qwen3-VL-4B   & \textbf{0.6125$\pm$0.003} & 0.5989$\pm$0.003 \\
        \midrule
        \multirow{2}{*}{BERTScore-F1}
          & Qwen2.5-VL-3B & \textbf{0.7211$\pm$0.002} & 0.6952$\pm$0.002 \\
          & Qwen3-VL-4B   & \textbf{0.7193$\pm$0.003} & 0.7000$\pm$0.002 \\
        \midrule
        \multirow{2}{*}{ROUGE-L}
          & Qwen2.5-VL-3B & \textbf{0.3204$\pm$0.004} & 0.2792$\pm$0.004 \\
          & Qwen3-VL-4B   & \textbf{0.3203$\pm$0.005} & 0.2818$\pm$0.005 \\
        \bottomrule
    \end{tabular}%
    }
\end{table}

\section{Discussion}

We presented SemEnrich, which turns the structure of a radiology corpus into a self-supervised training signal: clustering report sentences and reinserting plausibly omitted normal observations enriches the training data without any external annotation, and consistently improves report generation across five LLM backbones. The same clustering is also reusable as a reward signal for GRPO post-training, where a cluster-F1 reward complements the exact-match reward and yields further gains. Enrichment and the reward are orthogonal -- one changes the training data by adding plausibly omitted normal sentences, while the other leaves the data untouched and instead reshapes the supervision signal at the cluster level. 

\paragraph{Limitations.}
The sentences added through our enrichment method are not clinically verified by domain experts. Although the enrichment is constrained by connectivity and polarity conditions---ensuring that only \emph{normal} findings from co-occurring clusters are appended---the added sentences are derived algorithmically from the training corpus rather than confirmed by a radiologist for each individual patient. As a result, the enriched reports should not be treated as ground-truth clinical documents; they are intended exclusively as a training signal for machine learning models, and in a clinical deployment setting all generated reports would still require review and validation by qualified radiologists before any diagnostic use. Our approach also inherits the limitations of the underlying clustering and sign-assignment steps: the choice of sentence encoder, clustering algorithm, and number of clusters affects both the granularity of the enrichment and the noise in cluster membership, and our sign labels rely on an LLM judgment on a single representative sentence per cluster rather than a curated taxonomy.



\clearpage
\bibliography{refs}

\appendix
\onecolumn
\section{Appendix}
\subsection{Cluster Examples}\label{sec:appendix-cluster-examples}
Additional semantic clusters, continuing from Cluster~1 in the main text.
\begin{tcolorbox}[
  colback=hueA!6,
  colframe=hueA!55,
  boxrule=0.4pt,
  arc=2pt,
  left=4pt, right=4pt, top=3pt, bottom=3pt,
  title={\small\textbf{Cluster 2:} IVC Filter},
  fonttitle=\small
]
\small\textit{``IVC filter noted within abdomen.'' / ``IVC filter seen in the abdomen.'' / ``IVC filter partially visualized in the abdomen.'' / ...}
\end{tcolorbox}\vspace{2pt}
\begin{tcolorbox}[
  colback=hueA!6,
  colframe=hueA!55,
  boxrule=0.4pt,
  arc=2pt,
  left=4pt, right=4pt, top=3pt, bottom=3pt,
  title={\small\textbf{Cluster 3:} Normal Size},
  fonttitle=\small
]
\small\textit{``moderate size is normal.'' / ``there is normal size.'' / ``it is grossly normal in size.'' / ``it is normal size and shape.'' / ...}
\end{tcolorbox}\vspace{2pt}
\begin{tcolorbox}[
  colback=hueA!6,
  colframe=hueA!55,
  boxrule=0.4pt,
  arc=2pt,
  left=4pt, right=4pt, top=3pt, bottom=3pt,
  title={\small\textbf{Cluster 4:} Imaging Artifacts},
  fonttitle=\small
]
\small\textit{``this could be due to the artifacts.'' / ``this may be an artifact.'' / ``this could be an artifact.'' / ``i suspect this is an artifact, but needs to be followed.'' / ...}
\end{tcolorbox}\vspace{2pt}
\begin{tcolorbox}[
    colback=hueA!6,
    colframe=hueA!55,
    boxrule=0.4pt,
    arc=2pt,
    left=4pt, right=4pt, top=3pt, bottom=3pt,
    title={\small\textbf{Cluster 5:} Pelvic Phleboliths},
    fonttitle=\small
  ]
  \small\textit{``bilateral pelvic phleboliths.'' / ``pelvic calcifications consistent with phleboliths.'' / ``multiple pelvic phleboliths.'' / ``numerous pelvic phleboliths are noted.'' / ...}
  \end{tcolorbox}\vspace{2pt}
  \begin{tcolorbox}[
    colback=hueA!6,
    colframe=hueA!55,
    boxrule=0.4pt,
    arc=2pt,
    left=4pt, right=4pt, top=3pt, bottom=3pt,
    title={\small\textbf{Cluster 6:} Cerebral Infarcts},
    fonttitle=\small
  ]
  \small\textit{``focal encephalomalacia within the right occipital lobe...consistent with remote infarcts.'' / ``there is a large left middle cerebral artery territory infarct...'' / ``there is an old left periventricular frontal lobe infarct.'' / ...}
  \end{tcolorbox}\vspace{2pt}
\begin{tcolorbox}[
  colback=hueA!6,
  colframe=hueA!55,
  boxrule=0.4pt,
  arc=2pt,
  left=4pt, right=4pt, top=3pt, bottom=3pt,
  title={\small\textbf{Cluster 7:} Sickle Cell Disease Changes},
  fonttitle=\small
]
\small\textit{``osseous changes of sickle cell disease are chronic and stable.'' / ``bony changes of sickle cell disease are noted.'' / ``biconcave vertebral body configuration is typical for sickle cell disease.'' / ...}
\end{tcolorbox}

\subsection{Enrichment Examples}\label{sec:appendix-enrichment-examples}
More examples of valid enrichments. We add a random sentence from an expansion cluster to the current sample during training. The first two boxes are alternative enrichments for the same initial findings used in the main-text example (Right hilar mass; heart normal; lungs clear; no pneumothorax or effusion; right-upper-lobe airspace disease), illustrating that one report can admit several distinct valid enrichments.

\begin{tcolorbox}[
  colback=hueF!7,
  colframe=hueF!60,
  boxrule=0.5pt,
  arc=2pt,
  title={\textbf{Alternative Enrichment A (Same Initial Findings)}},
  fonttitle=\small\bfseries
]
\textbf{Expansion Cluster 1:} \textit{``no bony abnormality is seen'', ``no acute bony abnormalities'', ``no acute bony abnormality is seen'', ...}\\
\textbf{Expansion Cluster 2:} \textit{``normal lung volumes'', ``lung volumes are normal'', ``the lung volumes are normal'', ...}
\end{tcolorbox}\vspace{2pt}
\begin{tcolorbox}[
  colback=hueF!7,
  colframe=hueF!60,
  boxrule=0.5pt,
  arc=2pt,
  title={\textbf{Alternative Enrichment B (Same Initial Findings)}},
  fonttitle=\small\bfseries
]
\textbf{Expansion Cluster 1:} \textit{``trachea is midline'', ``the trachea is midline'', ``midline trachea'', ...}\\
\textbf{Expansion Cluster 2:} \textit{``visualized bony thorax intact'', ``the visualized bony structures of the thorax are intact'', ...}
\end{tcolorbox}\vspace{2pt}
\begin{tcolorbox}[
  colback=hueF!7,
  colframe=hueF!60,
  boxrule=0.5pt,
  arc=2pt,
  title={\textbf{Enrichment Example 1}},
  fonttitle=\small\bfseries
]
\textbf{Initial Findings:}
\begin{itemize}[noitemsep,topsep=2pt,leftmargin=*]
  \item The heart size and mediastinal contours are within normal limits.
  \item Both lungs are clear.
  \item The visualized skeletal structures are unremarkable.
  \item Bilateral nipple rings.
\end{itemize}
\textbf{Expansion Cluster:} \textit{``there is no pleural effusion or edema identified'', ``no pleural effusions'', ``no pleural effusion'', ...}
\end{tcolorbox}\vspace{2pt}
\begin{tcolorbox}[
    colback=hueF!7,
    colframe=hueF!60,
    boxrule=0.5pt,
    arc=2pt,
    title={\textbf{Enrichment Example 2}},
    fonttitle=\small\bfseries
  ]
  \textbf{Initial Findings:}
  \begin{itemize}[noitemsep,topsep=2pt,leftmargin=*]
    \item Heart: Normal. Lungs: The lungs are hyperinflated.
    \item There is mild scattered chronic lung changes throughout both lung fields, most pronounced in the right apex.
    \item There is a 8 mm calcified granuloma in the left lung base. Otherwise, the lung fields are relatively clear.
    \item Mediastinum: Normal. Pleural effusions: None.
    \item There is suggestion of diffuse osteopenia. Correlate for osteoporosis.
    \item There is moderate-to-marked chronic anterior wedge compression deformity of the approximately T9 vertebral body.
  \end{itemize}
  \textbf{Expansion Cluster:} \textit{``no pneumothorax is seen'', ``there is no pneumothorax'', ``no definite pneumothorax is seen'', ...}
  \end{tcolorbox}

\newpage
\subsection{HDBSCAN Clustering Statistics}\label{sec:appendix-hdbscan-stats}

Although we have 184,535 sentences in the training dataset, as HDBSCAN allows outliers and not to assign them to a cluster,
eventuall we end up with 66,477 sentences in the training dataset that are assigned to a cluster.  This is one of the main difference between density based clustering and K-means.
\begin{table}[H]
  \centering
  \small
  \begin{tabular}{lc}
  \toprule
  \textbf{Statistic} & \textbf{Value} \\
  \midrule
  Total Sentences & 66,477 \\
  Total Clusters & 3,216 \\
  Average Cluster Size & 20.67 \\
  Median Cluster Size & 11.0 \\
  Min Cluster Size & 5 \\
  Max Cluster Size & 1,522 \\
  \bottomrule
  \end{tabular}
  \caption{HDBSCAN clustering statistics for finding sentences.}
  \label{tab:hdbscan-stats-appendix}
\end{table}\vspace{-6pt}
\begin{figure}[H]
  \centering
  \begin{subfigure}[b]{0.32\columnwidth}
    \includegraphics[width=\linewidth]{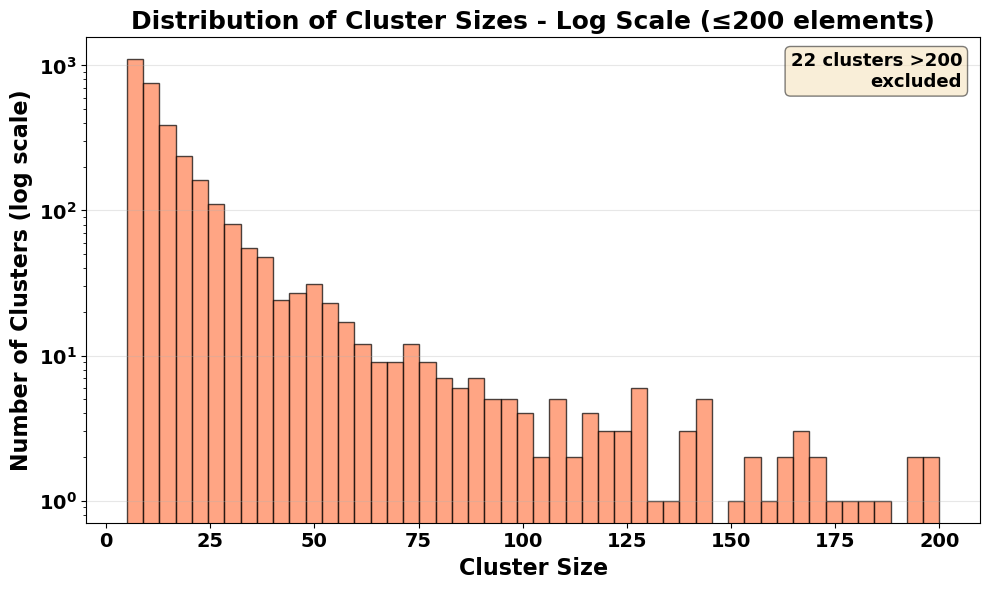}
    \caption{}
    \label{fig:hdbscan-log-histogram-appendix}
  \end{subfigure}
  \hfill
  \begin{subfigure}[b]{0.32\columnwidth}
    \includegraphics[width=\linewidth]{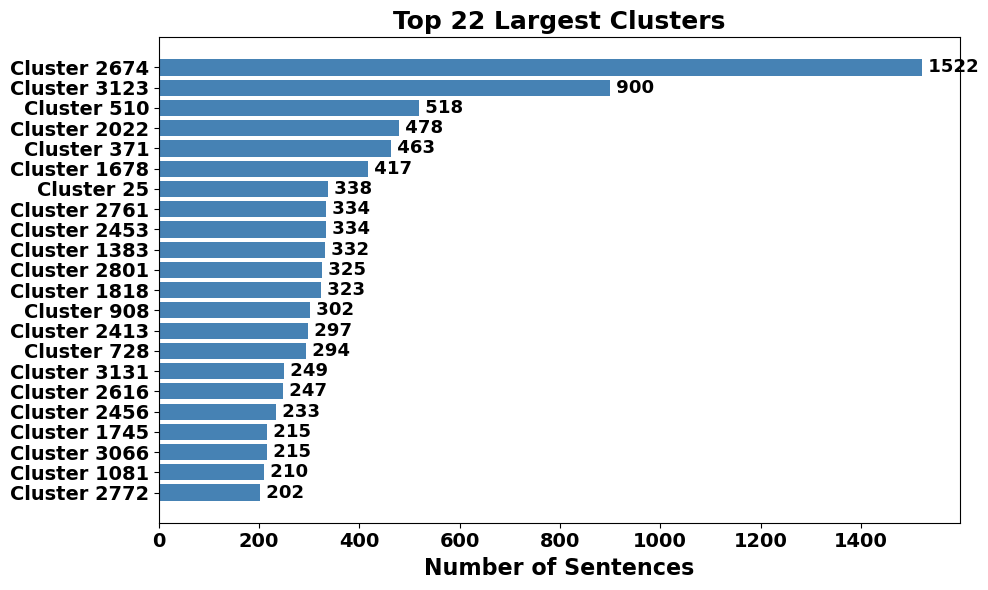}
    \caption{}
    \label{fig:hdbscan-top-clusters}
  \end{subfigure}
  \hfill
  \begin{subfigure}[b]{0.32\columnwidth}
    \includegraphics[width=\linewidth]{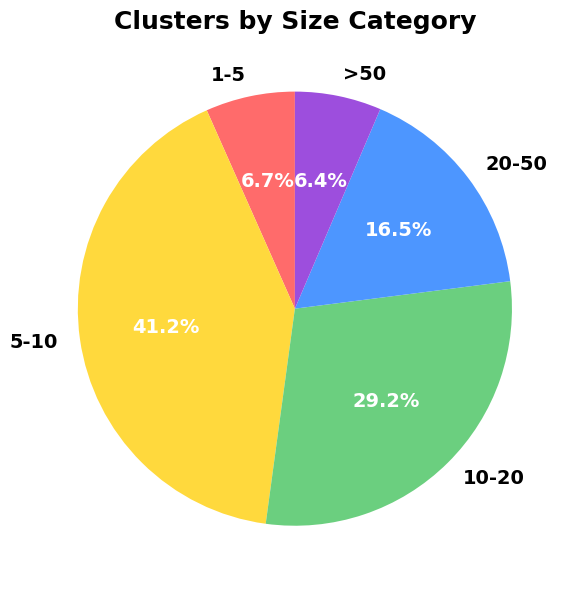}
    \caption{}
    \label{fig:hdbscan-percentages}
  \end{subfigure}
  \caption{HDBSCAN clustering statistics: (a) Log histogram of cluster sizes, (b) Top clusters by size, (c) Percentage distribution of clusters.}
  \label{fig:hdbscan-stats-appendix}
\end{figure}




\subsection{K-means Clustering Statistics with 5000 Clusters}\label{sec:appendix-kmeans-stats}
We provide the details of the K-means clustering with 5000 clusters as an example to illustrate the clustering difference between density based clustering methods and K-means.
\begin{table}[H]
  \centering
  \small
  \begin{tabular}{lc}
  \toprule
  \textbf{Statistic} & \textbf{Value} \\
  \midrule
  Total Sentences & 184,535 \\
  Total Clusters & 5,000 \\
  Average Cluster Size & 36.91 \\
  Median Cluster Size & 31.0 \\
  Min Cluster Size & 1 \\
  Max Cluster Size & 296 \\
  \bottomrule
  \end{tabular}
  \caption{K-means-5000 clustering statistics for finding sentences.}
  \label{tab:kmeans-5000-stats}
\end{table}\vspace{-6pt}
\begin{figure}[H]
  \centering
  \begin{subfigure}[b]{0.32\columnwidth}
    \includegraphics[width=\linewidth]{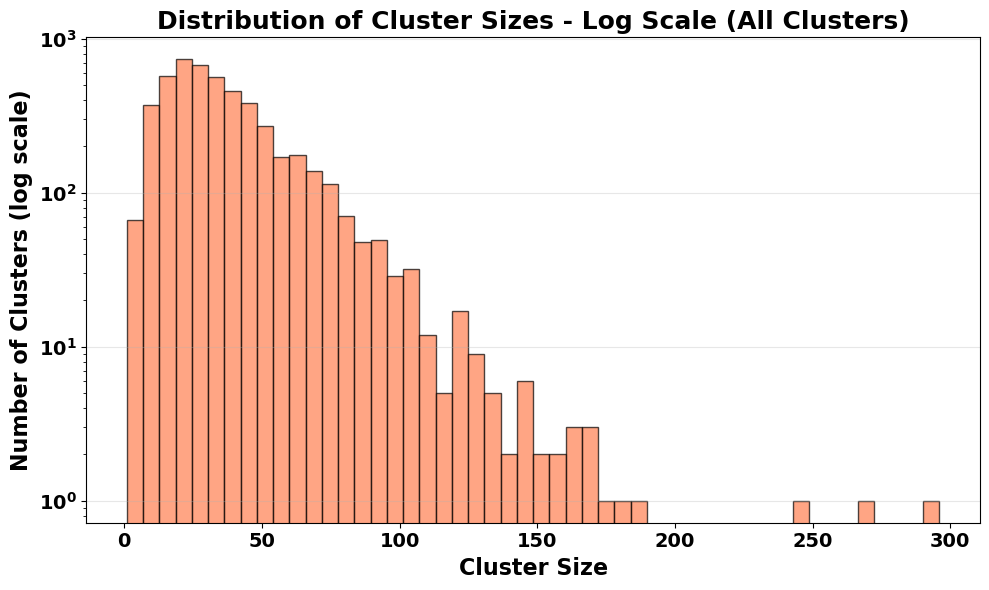}
    \caption{}
    \label{fig:kmeans_5000-log-histogram}
  \end{subfigure}
  \hfill
  \begin{subfigure}[b]{0.32\columnwidth}
    \includegraphics[width=\linewidth]{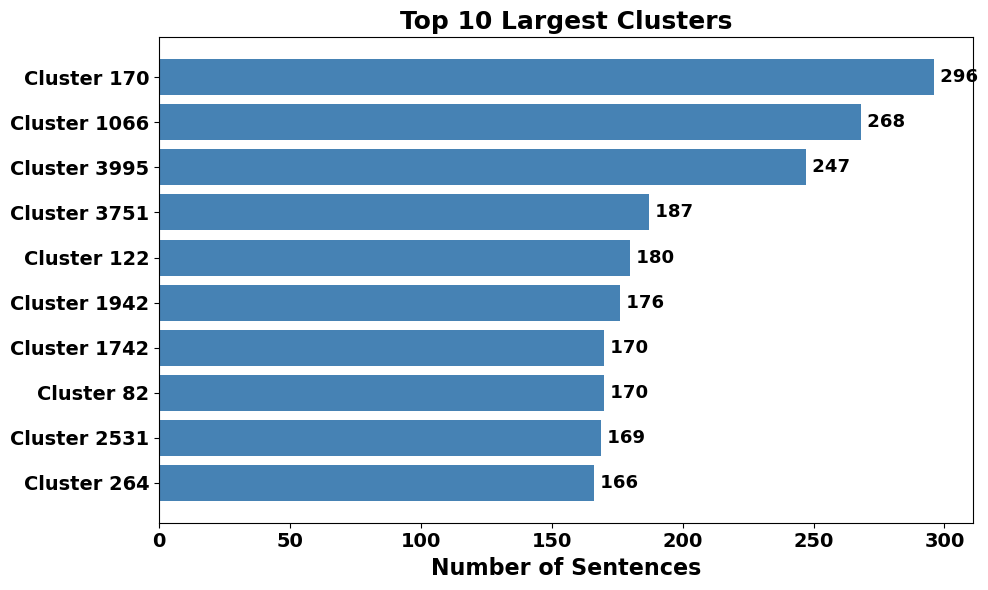}
    \caption{}
    \label{fig:kmeans_5000-top-clusters}
  \end{subfigure}
  \hfill
  \begin{subfigure}[b]{0.32\columnwidth}
    \includegraphics[width=\linewidth]{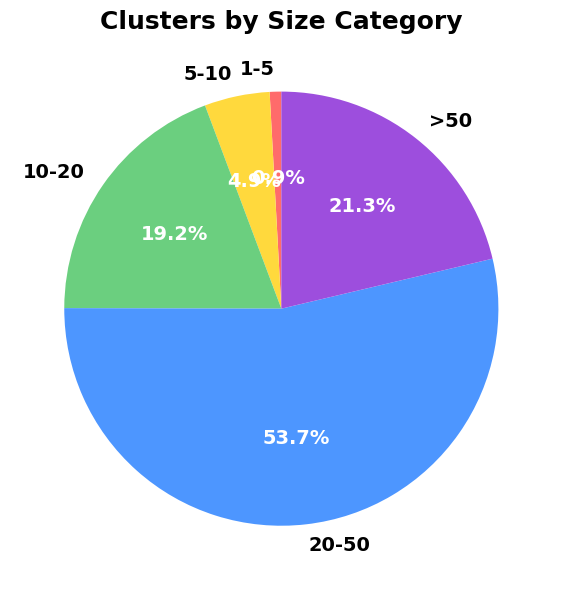}
    \caption{}
    \label{fig:kmeans_5000-percentages}
  \end{subfigure}
  \caption{K-means-5000 clustering statistics: (a) Log histogram of cluster sizes, (b) Top clusters by size, (c) Percentage distribution of clusters.}
  \label{fig:kmeans_5000-stats}
\end{figure}

\newpage
\subsection{Largest Valid Enrichment Problem}\label{sec:maximal_cliques_problem}

The problem of finding all maximal expansions in our framework is closely 
related to the classical \emph{maximal clique enumeration} problem in graph 
theory. Given an undirected graph $G = (V, E)$, a \emph{clique} is a subset 
of vertices in which every pair is connected by an edge. A clique is 
\emph{maximal} if no additional vertex can be included without breaking the 
clique property, and it is \emph{maximum} if it has the largest cardinality 
among all cliques in the graph. The decision version of the maximum clique 
problem---determining whether a clique of size at least $k$ exists---is one 
of Karp's original 21 NP-complete problems~\cite{karp2009reducibility}. 
Moon and Moser~\cite{moon1965cliques} established that a graph on $n$ 
vertices can contain up to $3^{n/3}$ maximal cliques, and this bound is 
tight. Consequently, any algorithm that enumerates all maximal cliques 
requires exponential time in the worst case simply because the output 
itself can be exponentially large.

The most widely used algorithm for maximal clique enumeration is the 
Bron--Kerbosch algorithm~\cite{bron1973algorithm}, a backtracking 
procedure that maintains three sets: $R$ (the current clique), $P$ 
(candidate vertices that can extend $R$), and $X$ (vertices already 
processed that ensure maximality). With the \emph{pivoting} 
optimization~\cite{tomita2006worst}, the algorithm achieves 
$O(3^{n/3})$ worst-case time, matching the Moon--Moser bound and 
making it output-optimal. Furthermore, 
Tsukiyama et al.~\cite{tsukiyama1977new} showed that maximal cliques 
can be enumerated with \emph{polynomial delay}---the time between 
outputting consecutive cliques is bounded by a polynomial in $n$.

In our setting, finding all largest valid enrichments for a patient with 
findings $F_j$ corresponds to enumerating all maximal cliques in the 
subgraph induced by $\mathcal{N}^+(F_j)$: the set of normal clusters
that are addable to every cluster in $F_j$. Since $|\mathcal{N}^+(F_j)|$
is typically much smaller than the total number of clusters $|C|$---due to
the intersection of neighborhoods---the enumeration remains tractable in
practice. We precompute all maximal expansions for every unique finding
graph before training, and during training we uniformly sample one
expansion per instance in each batch.

The addability matrix $B$ in Algorithm~\ref{alg:expansion} depends on two
thresholds, which we fix across all configurations: $\tau_{\text{norm}}=0.001$,
a floor on the row-normalized co-occurrence $\tilde{A}[i,j]$ between two
clusters, and $\tau_{\text{count}}=3$, a floor on the raw number of reports
$A[i,j]$ in which they co-occur. A candidate normal cluster is addable to a
report only if it clears both thresholds with respect to every cluster
already present, in addition to passing the polarity gate ($h_S(j)=+1$).

%
\newpage
\begin{algorithm}[H]
    \caption{Precomputing All Maximal Valid Enrichments}
    \label{alg:expansion}
    \begin{algorithmic}[1]
    \REQUIRE Training data $\mathcal{D} = \{(v_i, F_i)\}_{i=1}^{N}$, cluster mapping $f_C$, cluster signs $h_S$, thresholds $\tau_{\text{norm}}, \tau_{\text{count}}$
    \ENSURE Expansion dictionary $\mathcal{E}$
    
    \STATE \textbf{// Stage 1: Build Co-occurrence and Addability}
    \STATE Initialize $A \in \mathbb{R}^{K \times K} \leftarrow \mathbf{0}$ \hfill $\triangleright$ $K$ = number of clusters
    \FOR{each $(v_i, F_i) \in \mathcal{D}$}
        \STATE $G_i \leftarrow \{f_C(s) : s \in F_i\}$
        \FOR{each pair $(c_a, c_b) \in G_i \times G_i$, $c_a \neq c_b$}
            \STATE $A[c_a, c_b] \leftarrow A[c_a, c_b] + 1$
        \ENDFOR
    \ENDFOR
    \STATE $\tilde{A} \leftarrow \text{RowNormalize}(A)$
    \STATE $B[i,j] \leftarrow \mathbf{1}\!\left[\tilde{A}[i,j] > \tau_{\text{norm}} \;\wedge\; A[i,j] > \tau_{\text{count}} \;\wedge\; h_S(j) = +1\right]$
    
    \STATE
    \STATE \textbf{// Stage 2: Enumerate Maximal Expansions}
    \FOR{each unique finding graph $G$}
        \STATE $\mathcal{N}^+(G) \leftarrow \bigcap_{c \in G} \{j : B[c,j]=1\} \setminus G$ \hfill $\triangleright$ Compatible normal clusters
        \STATE $\mathcal{E}[G] \leftarrow \textsc{Expand}(G,\; \text{sorted}(\mathcal{N}^+(G)),\; 0)$
        \STATE Remove any $R_i \in \mathcal{E}[G]$ s.t.\ $\exists\, R_j \in \mathcal{E}[G]$: $R_i \subset R_j$ \hfill $\triangleright$ Keep only maximal
    \ENDFOR
    \RETURN $\mathcal{E}$
    
    \STATE
    \FUNCTION{\textsc{Expand}($R,\; P,\; s$)}
        \STATE $\text{expanded} \leftarrow \text{false}$
        \FOR{$i = s$ \TO $|P|$}
            \IF{$B[c,\, p_i] = 1$ for all $c \in R$}
                \STATE \textsc{Expand}($R \cup \{p_i\},\; P,\; i+1$)
                \STATE $\text{expanded} \leftarrow \text{true}$
            \ENDIF
        \ENDFOR
        \IF{\NOT $\text{expanded}$}
            \STATE \textbf{emit} $R$ \hfill $\triangleright$ No candidate can extend $R$
        \ENDIF
    \ENDFUNCTION
    \end{algorithmic}
    \end{algorithm}

\newpage
\subsection{Comparison with State-of-the-Art Medical VLMs}\label{sec:appendix-sota}

We benchmark our enriched Mistral-7B backbone against five open-source medical vision--language models on the ReXGradient-160K validation split ($2{,}500$ reports): Med-Flamingo \cite{moor2023med}, LLaVA-Med \cite{li2023llava}, CXR-LLaVA \cite{lee2025cxr}, CheXagent \cite{chen2024chexagent}, and LLaVA-Rad \cite{chaves2024training}. SOTA outputs are produced by each model's released checkpoint and evaluated with the same metric pipeline used for our backbones (in particular, RadGraph-F1 uses the \emph{complete}-match variant for both our methods and SOTA, matching the value reported for Mistral-7B in Table~\ref{tab:main_results}). SOTA numbers are therefore single inference runs (no seed variance), while our Mistral-7B rows are mean$\pm$std over five training seeds and are reproduced from Table~\ref{tab:main_results}.

We stress that this is \emph{not} a matched comparison and is included only to situate our absolute numbers: the SOTA models are evaluated zero-shot from released checkpoints and were never trained on ReXGradient-160K, whereas our backbone is fine-tuned in-domain. The table therefore mostly reflects the benefit of in-domain report-only fine-tuning, not the enrichment; the controlled evidence for enrichment is the baseline-vs-enriched comparison (Table~\ref{tab:main_results}) and the random-expansion ablation (Table~\ref{tab:ablation_results}).

Table~\ref{tab:sota_comparison} reports the four shared metrics. Our HDBSCAN-enriched Mistral-7B is the top model on all four: COMET ($+1.8$ over the strongest SOTA), BERTScore-F1 ($+5.6$), Sentence-BLEU-1 ($+14.1$), and RadGraph-F1 ($+16.5$). Indeed the unenriched Mistral-7B baseline already matches or exceeds every SOTA model on each metric, which confirms that the gap chiefly reflects in-domain fine-tuning; semantic enrichment then improves further on top of that already-strong baseline.

\begin{table}[t]
  \centering
  \caption{Comparison with open-source state-of-the-art medical VLMs on the ReXGradient-160K validation split ($2{,}500$ reports). SOTA rows are single inference runs produced by each model's released checkpoint; Mistral-7B rows are mean$\pm$std over five training seeds (taken from Table~\ref{tab:main_results}). RadGraph-F1 is the \emph{complete}-match variant for all rows. Best value per metric is bolded. All scores are in percent.}
  \label{tab:sota_comparison}
  \small
  \begin{tabular}{lcccc}
    \toprule
    Model & COMET & BERTScore & S-BLEU-1 & RadGraph \\
    \midrule
    Med-Flamingo & 51.18 & 62.63 & 6.79 & 10.47 \\
    LLaVA-Med & 61.31 & 58.17 & 8.80 & 4.44 \\
    CXR-LLaVA & 67.40 & 63.56 & 17.60 & 13.63 \\
    CheXagent & 60.81 & 63.46 & 17.66 & 14.83 \\
    LLaVA-Rad & 65.75 & 65.72 & 18.57 & 15.75 \\
    \midrule
    Mistral-7B (Baseline) & 63.59$\pm$0.77 & 67.86$\pm$0.75 & 29.87$\pm$1.67 & 30.21$\pm$1.72 \\
    Mistral-7B (HDBSCAN, ours) & \textbf{69.16$\pm$0.57} & \textbf{71.35$\pm$0.56} & \textbf{32.64$\pm$1.00} & \textbf{32.23$\pm$1.60} \\
    \bottomrule
  \end{tabular}
\end{table}

\newpage
\subsection{Generalization to MIMIC-CXR}\label{sec:appendix-mimic}

To assess whether SemEnrich transfers beyond ReXGradient-160K, we run the
complete pipeline---sentence embedding, clustering, polarity labelling,
connectivity, and enrichment---on the MIMIC-CXR dataset \cite{johnson2019mimic},
using a $20$K-report training subset. The pipeline transfers without
modification: HDBSCAN produces $836$ clusters, and enrichment touches $58.8\%$
of reports, adding on average $1.2$ sentences per enriched report. For polarity
labelling on MIMIC we use a $3$-sample majority vote per cluster to reduce the
single-sentence labelling noise discussed in the main text. We then fine-tune a
Mistral-7B backbone on the raw and enriched data under the same recipe as our
main experiments and report the mean$\pm$std over four seeds.

As shown in Table~\ref{tab:mimic_generalization}, HDBSCAN enrichment improves
the backbone on every metric, including the clinically grounded CheXbert-14
($+3.10$) and RadGraph-F1 ($+2.59$). MIMIC is a single-institution corpus with
noisier reports than ReXGradient-160K (Appendix~\ref{sec:dataset_appendix}), so
the fact that the enrichment still helps is evidence that it is not specific to
one dataset or reporting style.

\begin{table}[H]
  \centering
  \caption{Generalization of SemEnrich to MIMIC-CXR ($20$K training reports).
  Mistral-7B backbone, mean$\pm$std over four seeds; all scores in percent.
  $\Delta$ is HDBSCAN minus Baseline.}
  \label{tab:mimic_generalization}
  \small
  \begin{tabular}{lccc}
    \toprule
    Metric & Baseline & HDBSCAN (ours) & $\Delta$ \\
    \midrule
    Sentence-BLEU-1 & 23.88$\pm$0.91 & \textbf{30.35$\pm$0.18} & $+6.47$ \\
    METEOR          & 27.04$\pm$0.99 & \textbf{30.17$\pm$0.32} & $+3.13$ \\
    CheXbert-14     & 52.70$\pm$0.44 & \textbf{55.80$\pm$2.24} & $+3.10$ \\
    RadGraph-F1     & 22.69$\pm$1.17 & \textbf{25.28$\pm$0.72} & $+2.59$ \\
    \bottomrule
  \end{tabular}
\end{table}

\newpage
\subsection{Comparison with Medical VLMs on MIMIC-CXR}\label{sec:appendix-mimic-sota}

For completeness we also situate the MIMIC numbers against the same
open-source medical vision--language models used in
Appendix~\ref{sec:appendix-sota}. As before, this is \emph{not} a matched
comparison: the SOTA models are single zero-shot inference runs from released
checkpoints, and it is included only to locate our absolute performance. Note
that two of these models, CheXagent \cite{chen2024chexagent} and LLaVA-Rad
\cite{chaves2024training}, were themselves trained on MIMIC-CXR, so the setting
is if anything favourable to them.

Even so (Table~\ref{tab:mimic_sota}), our HDBSCAN-enriched Mistral-7B attains
the best RadGraph-F1 and the second-best CheXbert-14, and improves over its own
baseline on all four metrics. The lexical metrics (Sentence-BLEU-1, METEOR)
still favour the MIMIC-trained models, consistent with the domain match.

\begin{table}[H]
  \centering
  \caption{Comparison with open-source medical VLMs on MIMIC-CXR. SOTA rows are
  single inference runs from released checkpoints; our rows are the Mistral-7B
  means from Table~\ref{tab:mimic_generalization}. CheXagent and LLaVA-Rad were
  trained on MIMIC-CXR. Best value per metric is bolded; all scores in percent.}
  \label{tab:mimic_sota}
  \small
  \begin{tabular}{lcccc}
    \toprule
    Model & S-BLEU-1 & METEOR & CheXbert-14 & RadGraph \\
    \midrule
    Med-Flamingo \cite{moor2023med} & 13.05 & 13.62 & 54.00 & 18.59 \\
    LLaVA-Med \cite{li2023llava}    & 17.99 & 14.54 & 54.00 & 7.68 \\
    CXR-LLaVA \cite{lee2025cxr}     & 30.30 & 28.10 & 32.75 & 17.50 \\
    CheXagent \cite{chen2024chexagent} & 35.69 & 32.31 & \textbf{58.00} & 22.35 \\
    LLaVA-Rad \cite{chaves2024training} & \textbf{37.84} & \textbf{33.64} & 55.25 & 19.35 \\
    \midrule
    Ours (Baseline)         & 23.88 & 27.04 & 52.70 & 22.69 \\
    Ours (HDBSCAN)          & 30.35 & 30.17 & 55.80 & \textbf{25.28} \\
    \bottomrule
  \end{tabular}
\end{table}

\newpage
\subsection{Sentence-Encoder Ablation}\label{sec:appendix-encoder}

Since embedding quality determines cluster quality, which in turn determines
which sentences are eligible for enrichment, we ablate the sentence encoder
with the rest of the pipeline held fixed. We compare our default
\texttt{all-MiniLM-L6-v2} against two larger general-purpose encoders
(Stella, GTE-Qwen2) and two biomedical encoders (PubMedBERT, BioBERT). All
runs use a Mistral-7B backbone on MIMIC-CXR.

Table~\ref{tab:encoder_ablation} shows that \emph{every} encoder improves over
the baseline on all four metrics. The larger and biomedical encoders raise the
lexical scores (Sentence-BLEU-1, METEOR), but the general-purpose MiniLM
remains the strongest on both clinical metrics (CheXbert-14 and RadGraph-F1),
and the biomedical encoders do not beat it. Finally, the difference in cluster
coverage is not an embedding-quality artifact: the larger and biomedical
encoders assign \emph{fewer} sentences to clusters (e.g., MiniLM $16{,}961$ vs.\
PubMedBERT $6{,}855$ and BioBERT $8{,}179$), and sentences left unclustered
still remain in their reports---they are only excluded from the candidate
enrichment pool, not removed from training. We therefore keep MiniLM, also the
top-listed general-purpose model in the Sentence-Transformers library, as the
default.

\begin{table}[H]
  \centering
  \caption{Sentence-encoder ablation with the pipeline held fixed (Mistral-7B,
  MIMIC-CXR). ``Clust./Assig.'' is the number of clusters and the number of
  sentences assigned to a cluster. Best value per metric is bolded; scores in
  percent.}
  \label{tab:encoder_ablation}
  \small
  \begin{tabular}{lccccc}
    \toprule
    Encoder (dim) & S-BLEU-1 & METEOR & CheXbert-14 & RadGraph & Clust./Assig. \\
    \midrule
    Baseline            & 23.88 & 27.04 & 52.70 & 22.69 & --- \\
    MiniLM (384)        & 30.35 & 30.17 & \textbf{55.80} & \textbf{25.28} & 836 / 16{,}961 \\
    Stella (1024)       & 35.70 & 33.41 & 54.40 & 23.28 & 534 / 12{,}772 \\
    GTE-Qwen2 (1536)    & 33.76 & 31.97 & 52.90 & 22.05 & 576 / 11{,}399 \\
    PubMedBERT (768)    & \textbf{36.59} & \textbf{35.88} & 52.80 & 24.44 & 536 / 6{,}855 \\
    BioBERT (768)       & 24.75 & 27.70 & 54.10 & 23.13 & 628 / 8{,}179 \\
    \bottomrule
  \end{tabular}
\end{table}

\newpage
\subsection{Which Normal Findings Are Recovered}\label{sec:appendix-recovered-normals}

To show that the enrichment repairs a systematic reporting bias rather than
merely moving a metric, we inspect what the procedure actually inserts. On
MIMIC-CXR, of the $836$ clusters, $133$ are labelled normal and $703$ abnormal;
enrichment touches $58.8\%$ of reports ($+1.2$ sentences each), and by
construction every inserted cluster is polarity-normal. The normal findings
that are most frequently recovered (i.e.\ most often omitted from the original
report) are the canonical ones, as shown in Table~\ref{tab:recovered_normals}.

\begin{table}[H]
  \centering
  \caption{Most frequently recovered normal findings on MIMIC-CXR, as a
  percentage of training reports into which they are inserted.}
  \label{tab:recovered_normals}
  \small
  \begin{tabular}{lc}
    \toprule
    Recovered normal finding & \% of reports \\
    \midrule
    normal heart size                    & 22.0 \\
    no osseous abnormalities             & 21.8 \\
    no pleural effusion or pneumothorax  & 19.6 \\
    no pneumothorax                      & 11.0 \\
    normal pulmonary vasculature         & 8.2 \\
    \bottomrule
  \end{tabular}
\end{table}

Aligning baseline and enriched generations exposes the mechanism directly: the
enriched model adds omitted normals (e.g., ``hilar contours unremarkable''),
which helps when the reference report states them and lowers overlap only when
the reference is terse. In the latter case the drop reflects the metric
penalising completeness rather than a genuine error, consistent with the
enrichment making implicit normal findings explicit.

\newpage
\subsection{Within-Cluster Polarity Purity}\label{sec:appendix-purity}

Because only clusters labelled \emph{normal} are ever inserted, what matters for
label noise is the polarity purity of those clusters. We therefore LLM-labelled
$\sim\!1{,}000$ member sentences across all $133$ inserted normal clusters on
MIMIC-CXR (for large clusters we sample up to $10$ members) and measured how
consistently each cluster's members share its assigned polarity.
Table~\ref{tab:cluster_purity} shows the clusters are highly pure.

\begin{table}[H]
  \centering
  \caption{Polarity purity of the $133$ inserted normal clusters (MIMIC-CXR).
  Insertion-weighted purity weights each cluster by how often it is inserted
  during training; the abnormal-injection rate is its complement.}
  \label{tab:cluster_purity}
  \small
  \begin{tabular}{lc}
    \toprule
    Statistic & \% \\
    \midrule
    Mean purity                 & 97.9 \\
    Median purity               & 100.0 \\
    Clusters $100\%$ pure       & 92.5 \\
    Clusters $\ge 80\%$ pure    & 97.0 \\
    Insertion-weighted purity   & 99.78 \\
    Abnormal-injection rate     & 0.22 \\
    \bottomrule
  \end{tabular}
\end{table}

Weighting each cluster by how often it is actually inserted during training, the
clusters we add are $\sim\!99.8\%$ pure, corresponding to an effective
abnormal-injection rate of only $\sim\!0.22\%$; in other words, the procedure
essentially never introduces an abnormal finding. The few genuinely mixed
clusters are rare and low-frequency, so a conservative safeguard is nearly free:
dropping every cluster below $80\%$ purity would remove just $0.27\%$ of the
total insertion mass. As noted above, our MIMIC runs additionally use a
$3$-sample majority vote for the cluster sign to further reduce labelling noise.

\newpage
\subsection{Semantic Enrichment Additional Results}\label{sec:appendix-additional_results}
We first report the CheXbert-5 F1 block deferred from the main results (Table~\ref{tab:main_results_chexbert5}); the corresponding CheXbert-5 ablation block is reported in the Ablation Study Results section below.

\begin{table}[H]
  \caption{CheXbert-5 F1 scores comparing baseline with clustering methods (deferred from the main results).}
  \label{tab:main_results_chexbert5}
  \centering
  \resizebox{\columnwidth}{!}{%
        \begin{tabular}{lcccccc}
          \toprule
      Model & Baseline & DBSCAN & HDBSCAN & K-means-1000 & K-means-2000 & K-means-5000 \\
      \midrule
      DS-R1-Qwen-1.5B & 17.26$\pm$4.85 & 18.37$\pm$2.99 & 21.19$\pm$5.09 & \textbf{23.54$\pm$4.17} & 23.23$\pm$4.84 & 21.65$\pm$4.93 \\
      Gemma3-4B & 17.73$\pm$1.85 & 9.34$\pm$2.77 & \textbf{20.06$\pm$2.05} & 17.17$\pm$4.32 & 11.07$\pm$2.18 & 12.88$\pm$1.16 \\
      DS-R1-Llama-8B & \textbf{31.90$\pm$3.44} & 23.15$\pm$2.27 & 24.93$\pm$4.47 & 29.70$\pm$2.37 & 26.30$\pm$3.89 & 27.51$\pm$5.28 \\
      Qwen3-8B & 21.97$\pm$5.39 & 22.48$\pm$4.14 & 25.02$\pm$3.51 & \textbf{26.22$\pm$3.12} & 21.78$\pm$4.80 & 25.77$\pm$4.69 \\
      Mistral-7B & 15.89$\pm$4.30 & 17.01$\pm$4.22 & 19.13$\pm$5.79 & \textbf{24.82$\pm$3.76} & 16.03$\pm$3.01 & 21.67$\pm$2.87 \\
      \midrule
      \textit{Mean} & 20.95 & 18.07 & 22.06 & \textbf{24.29} & 19.68 & 21.90 \\
          \bottomrule
    \end{tabular}%
  }
\end{table}\vspace{-6pt}

We additionally provide enrichment results for METEOR, ChrF++, ROUGE, and BERTScore (using RoBERTa-large as the underlying encoder).

\begin{table}[H]
  \caption{METEOR scores comparing baseline with clustering methods}
  \label{tab:avg-METEOR_pretrain}
  \centering
  \resizebox{\columnwidth}{!}{%
        \begin{tabular}{lcccccc}
          \toprule
      Model & Baseline & DBSCAN & HDBSCAN & K-means-1000 & K-means-2000 & K-means-5000 \\
      \midrule
      DS-R1-Qwen-1.5B & 26.36$\pm$0.98 & 26.97$\pm$1.70 & 30.13$\pm$1.18 & 29.90$\pm$1.27 & \textbf{30.94$\pm$1.53} & 28.63$\pm$1.20 \\
      Gemma3-4B & 20.48$\pm$0.64 & 20.70$\pm$0.98 & 23.52$\pm$1.16 & \textbf{25.68$\pm$0.95} & 23.18$\pm$0.64 & 23.04$\pm$0.08 \\
      DS-R1-Llama-8B & 26.26$\pm$1.12 & 28.10$\pm$1.21 & 29.72$\pm$0.79 & \textbf{30.87$\pm$1.43} & 30.51$\pm$1.05 & 30.46$\pm$1.27 \\
      Qwen3-8B & 27.21$\pm$1.43 & 27.57$\pm$0.89 & 31.21$\pm$1.97 & 29.92$\pm$1.37 & \textbf{31.57$\pm$1.36} & 30.38$\pm$0.86 \\
      Mistral-7B & 29.42$\pm$1.83 & 27.87$\pm$1.52 & 33.51$\pm$1.66 & 30.67$\pm$1.09 & 33.13$\pm$1.41 & \textbf{33.78$\pm$1.63} \\
      \midrule
      \textit{Mean} & 25.95 & 26.24 & 29.62 & 29.41 & \textbf{29.87} & 29.26 \\
          \bottomrule
    \end{tabular}%
  }
\end{table}\vspace{-6pt}
\begin{table}[H]
  \caption{ChrF++ scores comparing baseline with clustering methods}
  \label{tab:avg-sentence-ChrF++_pretrain}
  \centering
  \resizebox{\columnwidth}{!}{%
        \begin{tabular}{lcccccc}
          \toprule
      Model & Baseline & DBSCAN & HDBSCAN & K-means-1000 & K-means-2000 & K-means-5000 \\
      \midrule
      DS-R1-Qwen-1.5B & 35.39$\pm$0.81 & 36.90$\pm$1.48 & 39.40$\pm$1.04 & 40.43$\pm$0.94 & \textbf{40.88$\pm$1.41} & 38.72$\pm$1.03 \\
      Gemma3-4B & 31.61$\pm$0.66 & 32.40$\pm$0.71 & 34.50$\pm$1.15 & \textbf{37.03$\pm$0.75} & 34.91$\pm$0.57 & 34.34$\pm$0.09 \\
      DS-R1-Llama-8B & 35.33$\pm$0.88 & 38.13$\pm$1.07 & 38.99$\pm$0.75 & \textbf{41.42$\pm$1.22} & 40.83$\pm$1.01 & 40.03$\pm$1.18 \\
      Qwen3-8B & 35.97$\pm$1.43 & 37.66$\pm$0.67 & 39.99$\pm$1.60 & 40.69$\pm$0.91 & \textbf{41.26$\pm$1.15} & 39.85$\pm$0.81 \\
      Mistral-7B & 37.48$\pm$1.65 & 38.85$\pm$1.41 & 42.67$\pm$1.42 & 41.83$\pm$0.77 & 43.23$\pm$1.32 & \textbf{43.25$\pm$1.50} \\
      \midrule
      \textit{Mean} & 35.16 & 36.79 & 39.11 & 40.28 & \textbf{40.22} & 39.24 \\
          \bottomrule
    \end{tabular}%
  }
\end{table}\vspace{-6pt}
\begin{table}[H]
  \caption{Corpus BLEU-1 scores comparing baseline with clustering methods (Pretrain).}
  \label{tab:corpus-BLEU-1_pretrain}
  \centering
  \resizebox{\columnwidth}{!}{%
        \begin{tabular}{lcccccc}
          \toprule
      Model & Baseline & DBSCAN & HDBSCAN & K-means-1000 & K-means-2000 & K-means-5000 \\
      \midrule
      DS-R1-Qwen-1.5B & 25.59$\pm$0.95 & 29.53$\pm$1.42 & \textbf{32.44$\pm$0.96} & 27.42$\pm$0.67 & 31.05$\pm$0.65 & 31.98$\pm$0.86 \\
      Gemma3-4B & 23.67$\pm$0.69 & 24.77$\pm$1.05 & \textbf{27.64$\pm$0.95} & 23.19$\pm$0.71 & 23.72$\pm$0.74 & 26.93$\pm$0.49 \\
      DS-R1-Llama-8B & 26.73$\pm$1.17 & 31.19$\pm$1.30 & 32.02$\pm$0.88 & 28.06$\pm$1.19 & 31.86$\pm$0.89 & \textbf{33.17$\pm$0.91} \\
      Qwen3-8B & 26.14$\pm$1.21 & 29.93$\pm$0.87 & 33.06$\pm$1.37 & 28.45$\pm$0.69 & 32.09$\pm$0.90 & \textbf{33.19$\pm$0.60} \\
      Mistral-7B & 25.44$\pm$1.51 & 29.89$\pm$1.02 & \textbf{34.72$\pm$1.29} & 23.76$\pm$0.57 & 32.30$\pm$0.74 & 34.71$\pm$1.57 \\
      \midrule
      \textit{Mean} & 25.51 & 29.06 & \textbf{31.97} & 26.18 & 30.21 & 31.99 \\
          \bottomrule
    \end{tabular}%
  }
\end{table}

\vspace{-6pt}
\begin{table}[H]
  \caption{ROUGE-1-F scores comparing baseline with clustering methods (Pretrain).}
  \label{tab:ROUGE-1_F_pretrain}
  \centering
  \resizebox{\columnwidth}{!}{%
        \begin{tabular}{lcccccc}
          \toprule
      Model & Baseline & DBSCAN & HDBSCAN & K-means-1000 & K-means-2000 & K-means-5000 \\
      \midrule
      DS-R1-Qwen-1.5B & 37.79$\pm$0.99 & 36.13$\pm$1.34 & \textbf{40.54$\pm$0.99} & 37.53$\pm$0.75 & 39.83$\pm$1.20 & 38.96$\pm$0.86 \\
      Gemma3-4B & 31.59$\pm$0.86 & 30.52$\pm$0.96 & \textbf{34.09$\pm$1.21} & 33.51$\pm$0.71 & 32.43$\pm$0.44 & 33.32$\pm$0.17 \\
      DS-R1-Llama-8B & 37.80$\pm$0.81 & 37.80$\pm$0.83 & \textbf{40.47$\pm$0.79} & 38.27$\pm$1.04 & 40.24$\pm$0.93 & 40.39$\pm$0.85 \\
      Qwen3-8B & 38.35$\pm$1.30 & 37.09$\pm$0.59 & \textbf{41.40$\pm$1.52} & 38.10$\pm$0.69 & 40.36$\pm$0.99 & 40.42$\pm$0.93 \\
      Mistral-7B & 39.44$\pm$1.68 & 36.86$\pm$1.19 & \textbf{44.12$\pm$1.15} & 34.75$\pm$0.62 & 40.53$\pm$0.96 & 43.86$\pm$1.37 \\
      \midrule
      \textit{Mean} & 37.00 & 35.68 & \textbf{40.12} & 36.43 & 38.68 & 39.39 \\
          \bottomrule
    \end{tabular}%
  }
\end{table}\vspace{-6pt}
\begin{table}[H]
  \caption{ROUGE-2-F scores comparing baseline with clustering methods (Pretrain).}
  \label{tab:ROUGE-2_F_pretrain}
  \centering
  \resizebox{\columnwidth}{!}{%
        \begin{tabular}{lcccccc}
          \toprule
      Model & Baseline & DBSCAN & HDBSCAN & K-means-1000 & K-means-2000 & K-means-5000 \\
      \midrule
      DS-R1-Qwen-1.5B & \textbf{19.04$\pm$0.86} & 17.45$\pm$1.57 & 19.04$\pm$1.09 & 15.83$\pm$1.00 & 18.48$\pm$1.44 & 17.21$\pm$0.96 \\
      Gemma3-4B & 12.06$\pm$0.51 & 10.05$\pm$0.75 & \textbf{12.82$\pm$0.87} & 11.55$\pm$0.68 & 10.55$\pm$0.68 & 11.87$\pm$0.41 \\
      DS-R1-Llama-8B & 18.76$\pm$0.84 & 18.39$\pm$1.06 & 18.58$\pm$0.96 & 16.63$\pm$1.33 & 18.24$\pm$1.02 & \textbf{18.96$\pm$1.04} \\
      Qwen3-8B & \textbf{20.27$\pm$1.74} & 18.22$\pm$0.65 & 20.10$\pm$1.75 & 16.29$\pm$1.03 & 18.99$\pm$1.38 & 18.74$\pm$0.99 \\
      Mistral-7B & \textbf{23.53$\pm$1.87} & 20.96$\pm$1.57 & 22.56$\pm$1.53 & 15.82$\pm$0.82 & 21.12$\pm$1.26 & 23.34$\pm$1.43 \\
      \midrule
      \textit{Mean} & \textbf{18.73} & 17.01 & 18.62 & 15.22 & 17.48 & 18.02 \\
          \bottomrule
    \end{tabular}%
  }
\end{table}\vspace{-6pt}
\begin{table}[H]
  \caption{ROUGE-L-F scores comparing baseline with clustering methods (Pretrain).}
  \label{tab:ROUGE-L_F_pretrain}
  \centering
  \resizebox{\columnwidth}{!}{%
        \begin{tabular}{lcccccc}
          \toprule
      Model & Baseline & DBSCAN & HDBSCAN & K-means-1000 & K-means-2000 & K-means-5000 \\
      \midrule
      DS-R1-Qwen-1.5B & 30.32$\pm$0.89 & 28.32$\pm$1.22 & \textbf{30.49$\pm$1.08} & 26.60$\pm$0.72 & 29.39$\pm$1.07 & 28.79$\pm$0.57 \\
      Gemma3-4B & 23.64$\pm$0.52 & 21.89$\pm$0.66 & \textbf{24.67$\pm$0.88} & 22.54$\pm$0.69 & 22.27$\pm$0.55 & 23.85$\pm$0.42 \\
      DS-R1-Llama-8B & 29.96$\pm$0.76 & 29.40$\pm$0.98 & 30.32$\pm$0.84 & 27.17$\pm$0.97 & 29.48$\pm$0.76 & \textbf{30.45$\pm$0.82} \\
      Qwen3-8B & 31.06$\pm$1.22 & 29.22$\pm$0.64 & \textbf{31.27$\pm$1.33} & 26.99$\pm$0.84 & 29.89$\pm$1.05 & 30.14$\pm$0.65 \\
      Mistral-7B & 33.43$\pm$1.51 & 30.70$\pm$1.21 & 33.65$\pm$1.23 & 24.56$\pm$0.56 & 31.30$\pm$1.01 & \textbf{34.26$\pm$1.15} \\
      \midrule
      \textit{Mean} & 29.68 & 27.91 & \textbf{30.08} & 25.57 & 28.47 & 29.50 \\
          \bottomrule
    \end{tabular}%
  }
\end{table}

\subsection{Ablation Study Results}

We provided additional results for the ablation study below. For some metrics, although we observe an improvement from the random expansion due to metric hacking, we also see that semantically guided expansion yields additional improvements over the random expansion.

We first report the CheXbert-5 F1 ablation block deferred from the main paper.

\begin{table}[H]
  \caption{Ablation study: CheXbert-5 F1 comparing Baseline, HDBSCAN, and Random expansion.}
  \label{tab:ablation_chexbert-5-appendix}
  \centering
  \small
  \begin{tabular}{lccc}
      \toprule
      Model & Baseline & HDBSCAN & Random expansion \\
      \midrule
      DS-R1-Qwen-1.5B & 17.26$\pm$4.85 & \textbf{21.19$\pm$5.09} & 20.86$\pm$5.00 \\
      Gemma3-4B & 17.73$\pm$1.85 & \textbf{20.06$\pm$2.05} & 18.98$\pm$5.45 \\
      DS-R1-Llama-8B & \textbf{31.90$\pm$3.44} & 24.93$\pm$4.47 & 21.68$\pm$4.60 \\
      Qwen3-8B & 21.97$\pm$5.39 & 25.02$\pm$3.51 & \textbf{26.82$\pm$5.30} \\
      Mistral-7B & 15.89$\pm$4.30 & \textbf{19.13$\pm$5.79} & 14.84$\pm$2.02 \\
      \midrule
      \textit{Mean} & 20.95 & \textbf{22.06} & 20.63 \\
      \bottomrule
  \end{tabular}
\end{table}

\begin{table}[H]
  \caption{Ablation study: COMET comparing Baseline, HDBSCAN, and Random Expansion.}
  \label{tab:ablation_comet-kiwi-appendix}
  \centering
  \small
  \begin{tabular}{lccc}
      \toprule
      Model & Baseline & HDBSCAN & Random expansion \\
      \midrule
      DS-R1-Qwen-1.5B & 63.68$\pm$0.49 & \textbf{67.34$\pm$0.41} & 63.97$\pm$0.48 \\
      Gemma3-4B & 62.05$\pm$0.53 & \textbf{64.59$\pm$0.65} & 61.97$\pm$0.41 \\
      DS-R1-Llama-8B & 63.95$\pm$0.48 & \textbf{67.44$\pm$0.31} & 64.53$\pm$0.57 \\
      Qwen3-8B & 63.80$\pm$0.80 & \textbf{67.71$\pm$0.73} & 64.35$\pm$0.28 \\
      Mistral-7B & 63.59$\pm$0.77 & \textbf{69.16$\pm$0.57} & 61.53$\pm$0.47 \\
      \midrule
      \textit{Mean} & 63.41 & \textbf{67.25} & 63.27 \\
      \bottomrule
  \end{tabular}
\end{table}

\begin{table}[H]
  \caption{Ablation study: Sentence BLEU-1 comparing Baseline, HDBSCAN, and Random Expansion.}
  \label{tab:ablation_avg-sentence-bleu-1-appendix}
  \centering
  \small
  \begin{tabular}{lccc}
      \toprule
      Model & Baseline & HDBSCAN & Random expansion \\
      \midrule
      DS-R1-Qwen-1.5B & 27.58$\pm$0.87 & \textbf{29.51$\pm$0.81} & 26.88$\pm$1.01 \\
      Gemma3-4B & 21.63$\pm$0.70 & \textbf{23.83$\pm$0.91} & 22.18$\pm$0.49 \\
      DS-R1-Llama-8B & 27.45$\pm$0.86 & \textbf{29.11$\pm$0.54} & 27.59$\pm$0.71 \\
      Qwen3-8B & 28.11$\pm$1.17 & \textbf{30.35$\pm$1.35} & 27.17$\pm$0.44 \\
      Mistral-7B & 29.87$\pm$1.67 & \textbf{32.64$\pm$1.00} & 28.85$\pm$1.02 \\
      \midrule
      \textit{Mean} & 26.93 & \textbf{29.09} & 26.53 \\
      \bottomrule
  \end{tabular}
\end{table}

\begin{table}[H]
  \caption{Ablation study: METEOR comparing Baseline, HDBSCAN, and Random Expansion.}
  \label{tab:ablation_avg-meteor}
  \centering
  \small
  \begin{tabular}{lccc}
      \toprule
      Model & Baseline & HDBSCAN & Random expansion \\
      \midrule
      DS-R1-Qwen-1.5B & 26.36$\pm$0.98 & \textbf{30.13$\pm$1.18} & 28.33$\pm$1.41 \\
      Gemma3-4B & 20.48$\pm$0.64 & \textbf{23.52$\pm$1.16} & 22.24$\pm$0.61 \\
      DS-R1-Llama-8B & 26.26$\pm$1.12 & \textbf{29.72$\pm$0.79} & 28.74$\pm$1.24 \\
      Qwen3-8B & 27.21$\pm$1.43 & \textbf{31.21$\pm$1.97} & 28.34$\pm$0.60 \\
      Mistral-7B & 29.42$\pm$1.83 & \textbf{33.51$\pm$1.66} & 32.33$\pm$1.72 \\
      \midrule
      \textit{Mean} & 25.95 & \textbf{29.62} & 28.00 \\
      \bottomrule
  \end{tabular}
\end{table}


\begin{table}[H]
  \caption{Ablation study: ROUGE-1-F comparing Baseline, HDBSCAN, and Random Expansion.}
  \label{tab:ablation_rouge-1-f}
  \centering
  \small
  \begin{tabular}{lccc}
      \toprule
      Model & Baseline & HDBSCAN & Random expansion \\
      \midrule
      DS-R1-Qwen-1.5B & 37.79$\pm$0.99 & \textbf{40.54$\pm$0.99} & 36.73$\pm$1.11 \\
      Gemma3-4B & 31.59$\pm$0.86 & \textbf{34.09$\pm$1.21} & 31.77$\pm$0.59 \\
      DS-R1-Llama-8B & 37.80$\pm$0.81 & \textbf{40.47$\pm$0.79} & 37.79$\pm$0.86 \\
      Qwen3-8B & 38.35$\pm$1.30 & \textbf{41.40$\pm$1.52} & 37.28$\pm$0.44 \\
      Mistral-7B & 39.44$\pm$1.68 & \textbf{44.12$\pm$1.15} & 38.24$\pm$1.12 \\
      \midrule
      \textit{Mean} & 37.00 & \textbf{40.12} & 36.36 \\
      \bottomrule
  \end{tabular}
\end{table}

\begin{table}[H]
  \caption{Ablation study: ROUGE-2-F comparing Baseline, HDBSCAN, and Random Expansion.}
  \label{tab:ablation_rouge-2-f}
  \centering
  \small
  \begin{tabular}{lccc}
      \toprule
      Model & Baseline & HDBSCAN & Random expansion \\
      \midrule
      DS-R1-Qwen-1.5B & \textbf{19.04$\pm$0.86} & 19.04$\pm$1.09 & 17.42$\pm$1.20 \\
      Gemma3-4B & 12.06$\pm$0.51 & \textbf{12.82$\pm$0.87} & 11.95$\pm$0.81 \\
      DS-R1-Llama-8B & \textbf{18.76$\pm$0.84} & 18.58$\pm$0.96 & 18.21$\pm$1.32 \\
      Qwen3-8B & \textbf{20.27$\pm$1.74} & 20.10$\pm$1.75 & 17.82$\pm$0.45 \\
      Mistral-7B & \textbf{23.53$\pm$1.87} & 22.56$\pm$1.53 & 20.36$\pm$1.32 \\
      \midrule
      \textit{Mean} & \textbf{18.73} & 18.62 & 17.15 \\
      \bottomrule
  \end{tabular}
\end{table}

\begin{table}[H]
  \caption{Ablation study: ROUGE-L-F comparing Baseline, HDBSCAN, and Random Expansion.}
  \label{tab:ablation_rouge-l-f}
  \centering
  \small
  \begin{tabular}{lccc}
      \toprule
      Model & Baseline & HDBSCAN & Random expansion \\
      \midrule
      DS-R1-Qwen-1.5B & 30.32$\pm$0.89 & \textbf{30.49$\pm$1.08} & 28.04$\pm$0.89 \\
      Gemma3-4B & 23.64$\pm$0.52 & \textbf{24.67$\pm$0.88} & 23.28$\pm$0.70 \\
      DS-R1-Llama-8B & 29.96$\pm$0.76 & \textbf{30.32$\pm$0.84} & 29.19$\pm$0.95 \\
      Qwen3-8B & 31.06$\pm$1.22 & \textbf{31.27$\pm$1.33} & 28.55$\pm$0.42 \\
      Mistral-7B & 33.43$\pm$1.51 & \textbf{33.65$\pm$1.23} & 29.57$\pm$1.08 \\
      \midrule
      \textit{Mean} & 29.68 & \textbf{30.08} & 27.72 \\
      \bottomrule
  \end{tabular}
\end{table}

\begin{table}[H]
  \caption{Ablation study: ChrF++ comparing Baseline, HDBSCAN, and Random Expansion.}
  \label{tab:ablation_avg-sentence-chrf++}
  \centering
  \small
  \begin{tabular}{lccc}
      \toprule
      Model & Baseline & HDBSCAN & Random expansion \\
      \midrule
      DS-R1-Qwen-1.5B & 35.39$\pm$0.81 & \textbf{39.40$\pm$1.04} & 37.56$\pm$1.25 \\
      Gemma3-4B & 31.61$\pm$0.66 & \textbf{34.50$\pm$1.15} & 33.39$\pm$0.73 \\
      DS-R1-Llama-8B & 35.33$\pm$0.88 & \textbf{38.99$\pm$0.75} & 37.80$\pm$1.38 \\
      Qwen3-8B & 35.97$\pm$1.43 & \textbf{39.99$\pm$1.60} & 37.63$\pm$0.49 \\
      Mistral-7B & 37.48$\pm$1.65 & \textbf{42.67$\pm$1.42} & 40.02$\pm$1.46 \\
      \midrule
      \textit{Mean} & 35.16 & \textbf{39.11} & 37.28 \\
      \bottomrule
  \end{tabular}
\end{table}

\begin{table}[H]
  \caption{Ablation study: BERTScore-F1 comparing Baseline, HDBSCAN, and Random Expansion.}
  \label{tab:ablation_bertscore-f1-bert-base-uncased-appendix}
  \centering
  \small
  \begin{tabular}{lccc}
      \toprule
      Model & Baseline & HDBSCAN & Random expansion \\
      \midrule
      DS-R1-Qwen-1.5B & 67.60$\pm$0.48 & \textbf{69.66$\pm$0.42} & 67.06$\pm$0.55 \\
      Gemma3-4B & 65.35$\pm$0.31 & \textbf{66.80$\pm$0.60} & 65.43$\pm$0.31 \\
      DS-R1-Llama-8B & 67.83$\pm$0.44 & \textbf{69.80$\pm$0.34} & 67.91$\pm$0.55 \\
      Qwen3-8B & 67.79$\pm$0.67 & \textbf{70.07$\pm$0.69} & 67.62$\pm$0.11 \\
      Mistral-7B & 67.86$\pm$0.75 & \textbf{71.35$\pm$0.56} & 66.50$\pm$0.57 \\
      \midrule
      \textit{Mean} & 67.29 & \textbf{69.54} & 66.91 \\
      \bottomrule
  \end{tabular}
\end{table}

\subsection{GRPO Training Additional Results}
\label{sec:grpo_training_additional_results}
As we pointed out in the main paper, we observe significant performance improvements by using $\mathcal{R}_{cl}$ in the reward calculation of GRPO training. We provide results for different metrics below.

\begin{table}[H]
    \caption{Sentence-BLEU-1 scores, comparing $\mathcal{R}_{cl}$ and $\mathcal{R}_{exact}$ rewards (GRPO) on the impression section.}
    \label{tab:grpo_sentence_bleu1_impression}
    \centering
    \small
    \begin{tabular}{lcc}
        \toprule
        Model & $\mathcal{R}_{cl}$ & $\mathcal{R}_{exact}$ \\
        \midrule
        Qwen2.5-VL-3B & \textbf{0.2704$\pm$0.0035} & 0.2361$\pm$0.0042 \\
        Qwen3-VL-4B & \textbf{0.2706$\pm$0.0044} & 0.2354$\pm$0.0034 \\
        \bottomrule
    \end{tabular}
\end{table}

\begin{table}[H]
    \caption{METEOR scores, comparing $\mathcal{R}_{cl}$ and $\mathcal{R}_{exact}$ rewards (GRPO).}
    \label{tab:meteor_impression}
    \centering
    \resizebox{0.5\columnwidth}{!}{%
          \begin{tabular}{lcc}
            \toprule
        Model & $\mathcal{R}_{cl}$ & $\mathcal{R}_{exact}$ \\
        \midrule
        Qwen2.5-VL-3B & \textbf{0.2621 ± 0.0034} & 0.2335 ± 0.0054 \\
        Qwen3-VL-4B & \textbf{0.2620 ± 0.0048} & 0.2344 ± 0.0032 \\
            \bottomrule
      \end{tabular}%
    }
  \end{table}


\begin{table}[H]
    \caption{Sentence-BLEU-4 scores, comparing $\mathcal{R}_{cl}$ and $\mathcal{R}_{exact}$ rewards (GRPO).}
    \label{tab:sentence_bleu_4_impression}
    \centering
    \resizebox{0.5\columnwidth}{!}{%
          \begin{tabular}{lcc}
            \toprule
        Model & $\mathcal{R}_{cl}$ & $\mathcal{R}_{exact}$ \\
        \midrule
        Qwen2.5-VL-3B & \textbf{0.1825 ± 0.0029} & 0.1599 ± 0.0027 \\
        Qwen3-VL-4B & \textbf{0.1847 ± 0.0041} & 0.1578 ± 0.0023 \\
            \bottomrule
      \end{tabular}%
    }
  \end{table}

\subsection{On the Dataset}\label{sec:dataset_appendix}

As noted in the main text, we use the ReXGradient-160K dataset for our experiments \cite{zhang2025rexgradient}. 
This dataset comprises 160K chest X-ray studies covering 109,487 unique patients from 79 medical sites.
Since the data originates from 79 distinct sites, it exhibits considerable diversity in imaging conditions, acquisition protocols, and equipment.
Moreover, the reports are authored by different radiologists, introducing substantial variability in writing style and clinical language.

We did not use the CheXpert dataset \cite{irvin2019chexpert} for our experiments for two reasons. 
First, it is sourced solely from Stanford Hospital and therefore may lack the multi-site diversity present in ReXGradient-160K. 
Second, it is structured more as a classification benchmark than a free-text report generation dataset, 
which may not be representative of real-world clinical reporting scenarios.

We use the MIMIC-CXR dataset \cite{johnson2019mimic} not as our primary benchmark but as an out-of-distribution stress test of the enrichment (Appendix~\ref{sec:appendix-mimic}).
MIMIC is collected from a single medical center, which limits its institutional diversity relative to ReXGradient-160K, and its reports are noisier in the sense that findings or impressions may be missing or poorly structured---precisely the conditions under which our enrichment might be expected to struggle.
That the full pipeline nonetheless transfers and improves a backbone on MIMIC is therefore stronger evidence for its generality than a second clean corpus would be.

More broadly, single-institution datasets limit the evaluation of a model's generalization capability across diverse healthcare settings.
For these reasons, we chose the ReXGradient-160K dataset, which not only spans 79 medical centers but also ensures that every sample 
contains both findings and impressions, making it well suited for evaluating our method.

\subsection{On the Enriched Dataset}\label{sec:enrichment_limitation}

It is important to note that the sentences added through our enrichment method are not clinically verified by domain experts. 
Although the enrichment is constrained by connectivity and polarity conditions---ensuring that only \emph{normal} findings from co-occurring clusters are appended---the added sentences are derived algorithmically from the training corpus rather than confirmed by a radiologist for each individual patient.

As a result, the enriched reports should not be treated as ground-truth clinical documents. 
Instead, the enriched dataset is intended exclusively as a training signal for machine learning models. 
The goal is to provide the model with richer supervisory information during training by making implicit normal findings explicit, thereby encouraging the model to learn a more complete understanding of chest X-ray images. 

We emphasize that this enrichment strategy is a data augmentation technique for model training and is not a substitute for expert annotation. 
In a clinical deployment setting, all generated reports would still require review and validation by qualified radiologists before any diagnostic use.

\end{document}